\documentclass[9pt,a4paper,oneside]{extarticle} 

\usepackage{lineno} %

\usepackage[dvipsnames]{xcolor}
\usepackage{xspace}
\usepackage{graphicx}
\usepackage[left=1.2in, right=1.2in, top=0.9in, bottom=0.9in, a4paper]{geometry}
\usepackage[acronym]{glossaries}
\usepackage{authblk}
\usepackage{amssymb}
\usepackage{array}
\usepackage{multirow}
\usepackage{amssymb}
\usepackage{makecell}
\usepackage[sort&compress,numbers]{natbib}
\usepackage{pdflscape}
\usepackage{color, colortbl}
\usepackage{xr}

\definecolor{pastelred}{RGB}{255, 171, 171}
\definecolor{pastellightred}{RGB}{255, 203, 193}
\definecolor{pastelblue}{RGB}{110, 180, 255}
\definecolor{pastellightblue}{RGB}{133, 227, 255}
\definecolor{pastelgreen}{RGB}{175, 248, 219}
\definecolor{pastellightgreen}{RGB}{219, 255, 214}
\definecolor{pastelorange}{RGB}{244, 205, 166}
\definecolor{pastelgreen}{RGB}{152, 239, 171}
\definecolor{pastelviolet}{RGB}{194, 170, 230}

\newcolumntype{P}[1]{>{\centering\arraybackslash}p{#1}}
\newcolumntype{C}{>{\centering\arraybackslash}p{1.7em}}
\newcolumntype{D}{>{\centering\arraybackslash}p{3.4em}}
\newcolumntype{?}[1]{!{\vrule width #1}}
\newlength{\Oldarrayrulewidth}

\newacronym{abn}{ABN}{Attention Branch Network}
\newacronym{ai}{AI}{Artificial Intelligence}
\newacronym{ap}{AP}{Average Precision}
\newacronym{awd}{AWD}{Adaptive Wavelet Distillation}
\newacronym{cam}{CAM}{Class Activation Mapping}
\newacronym{cav}{CAV}{Concept Activation Vector}
\newacronym{cdep}{CDEP}{Contextual Decomposition Explanation Penalization}
\newacronym{ch}{CH}{Clever Hans}
\newacronym{clarc}{ClArC}{Class Artifact Compensation}
\newacronym{aclarc}{A-ClArC}{Augmentative Class Artifact Compensation}
\newacronym{pclarc}{P-ClArC}{Projective Class Artifact Compensation}
\newacronym{cnn}{CNN}{Convolutional Neural Network}
\newacronym{dnn}{DNN}{deep neural network}
\newacronym{dtd}{DTD}{Deep Taylor Decomposition}
\newacronym{ecq}{ECQ}{Entropy-constrained Quantization}
\newacronym{gan}{GAN}{Generative Adversarial Network}
\newacronym{gap}{GAP}{Global Average Pooling}
\newacronym{gcam}{Grad-CAM}{Gradient-weighted Class Activation Mapping}
\newacronym{ggcam}{Guided Grad-CAM}{Guided Gradient-weighted Class Activation Mapping}
\newacronym{spire}{SPIRE}{Spurious Pattern Identification and REpair}
\newacronym{ig}{IG}{Integrated Gradients}
\newacronym{lime}{LIME}{Local Interpretable Model-Agnostic Explanations}
\newacronym{lrp}{LRP}{Layer-wise Relevance Propagation}
\newacronym{ml}{ML}{machine learning}
\newacronym{nn}{NN}{neural network}
\newacronym{prp}{PRP}{Prototypical Relevance Propagation}
\newacronym{rrr}{RRR}{Right for the Right Reasons}
\newacronym{sgd}{SGD}{Stochastic Gradient Descent}
\newacronym{spray}{SpRAy}{Spectral Relevance Analysis}
\newacronym{svm}{SVM}{Support Vector Machine}
\newacronym{trim}{TRIM}{Transformation Importance}
\newacronym{vqa}{VQA}{Visual Question Answering}
\newacronym{xai}{XAI}{eXplainable Artificial Intelligence}
\newacronym{xil}{XIL}{eXplanatory Interactive Learning}
\newacronym{xpe}{XPE}{eXplanation-based performance estimation}

\newcommand{\ie}{{i.e.}\xspace}

\newcommand{\wrt}{{w.r.t.}\xspace}
\newcommand{\eg}{{e.g.}\xspace}

\begin{document}

\title{Beyond Explaining: \mbox{Opportunities and Challenges of XAI-Based Model Improvement}}

\renewcommand*{\thefootnote}{\fnsymbol{footnote}}
\author[1]{Leander~Weber}
\author[1]{Sebastian~Lapuschkin\footnote{corresponding: \texttt{\{sebastian.lapuschkin,wojciech.samek\}@hhi.fraunhofer.de}}}
\author[2, 4]{Alexander~Binder}
\newcommand\CorrAuthorMark{\footnotemark[\arabic{footnote}]}
\author[1, 3]{Wojciech~Samek\protect\CorrAuthorMark}
\affil[1]{\footnotesize Department of Artificial Intelligence, Fraunhofer Heinrich Hertz Institute, 10587 Berlin, Germany}
\affil[2]{\footnotesize ICT Cluster, Singapore Institute of Technology, 138683 Singapore, Singapore}
\affil[3]{\footnotesize BIFOLD – Berlin Institute for the Foundations of Learning and Data, Berlin, Germany}
\affil[4]{\footnotesize Department of Informatics, University of Oslo, 0373 Oslo, Norway}

\maketitle

\begin{abstract}
Explainable Artificial Intelligence (XAI) is an emerging research field bringing transparency to highly complex and opaque \glsfirst{ml} models. Despite the development of a multitude of methods to explain the decisions of black-box classifiers in recent years, these tools are seldomly used beyond visualization purposes. Only recently, researchers have started to employ explanations in practice to actually improve models. This paper offers a comprehensive overview over techniques that apply XAI practically for improving various properties of ML models, and systematically categorizes these approaches, comparing their respective strengths and weaknesses. We provide a theoretical perspective on these methods, and show empirically through experiments on toy and realistic settings how explanations can help improve properties such as model generalization ability or reasoning, among others. We further discuss potential caveats and drawbacks of these methods. We conclude that while model improvement based on XAI can have significant beneficial effects even on complex and not easily quantifyable model properties, these methods need to be applied carefully, since their success can vary depending on a multitude of factors, such as the model and dataset used, or the employed explanation method.
\end{abstract}

\section{Introduction}
\label{sec:intro}
In recent years, great advances have been made in the field of artificial intelligence, with especially \glspl{dnn} achieving impressive performances in a multitude of domains, from image classification (\cite{LeCun2015Deep, Krizhevsky2017Imagenet}) and the diagnosis of medical conditions (\cite{Kadampur2020Skin, Ali2020Smart}), to understanding chemical systems (\cite{Schuett2017Quantum, Wang2020Novel}), playing video games on a competitive level (\cite{Jaderberg2018Human, Vinyals2019Grandmaster}), predicting the weather (\cite{Shakya2020Deep, Hewage2021Deep}) or the spread of infectious diseases (\cite{Shahid2020Predictions, Arora2020Prediction}). And yet, the widespread application of these techniques in real-world scenarios has been hindered immensely by the black-box nature of \glspl{dnn}. The reasoning behind their decisions is generally not obvious, and as such, they are simply not trustworthy enough, as their decisions may be (and often are) biased, as shown in, \eg,  (\cite{Stock2018Convnets, Lapuschkin2019Unmasking, Anders2022Finding}).

In order to alleviate this problem, the field of \gls{xai} has recently shifted into focus, which aims to open the black box of deep learning models. This increases transparency and trust, by better understanding the reasoning of these models. In fact, a multitude of explanation methods has been developed, that are able to visualize the basis for a model's decision, all approaching the subject from different angles. Roughly, \textit{global} (\eg, \cite{Guyon2003Introduction, Kim2018Interpretability, Hu2020Architecture}) and \textit{local} (\eg, \cite{Baehrens2010How, Bach2015Pixel, Ribeiro2016Why}) \gls{xai} methods can be distinguished, with the former focusing on providing a general understanding of a model's learned concepts and internal representations, and the latter aiming to explain the reasoning behind specific, singular decisions. However, while all of the above methods offer various insights into a model's reasoning, the majority of research on the subject seems to stop here: Decisions are explained, and problems may be discovered, but the obtained insights are rarely applied to actually achieve \emph{more} trustworthy, fairer, or simply better performing models.

In contrast, the very similar idea of incorporating human knowledge into a \gls{ml} model's training for the purpose of correcting its reasoning is relatively old. By using expert knowledge and feedback to regularize the model's learning process, approaches such as \cite{Zaidan2007Using} in the context of Support Vector Machines managed to improve learning speed and reasoning as early as 2007. Very recently, there have been growing efforts to include explanations in a similar manner, with the aim of improving various properties of current \gls{ml} models. The techniques developed in this novel area of research are, however, conceptually extremely heterogeneous and their benefits and drawbacks are not well-studied. Also, to the best of our knowledge, so far there is no systematic review of \gls{xai}-based model improvement (although many reviews on explanation techniques have been recently published \cite{Samek2021Explaining, Tjoa2019Survey, Vilone2020Explainable}).

In this paper, we aim to close this gap by deriving an {\bf unifying theoretical framework} for XAI-based model improvement. We provide an exhaustive overview of the current state of this emerging research field by {\bf reviewing existing approaches} in a structured manner and {\bf evaluating them} systematically. Here, we consider factors such as the improved model property (performance, reasoning, equality, etc.), the augmented component, the degree of human involvement, and the model agnosticity. {\bf Showcasing various examples}, we further demonstrate the power of \gls{xai} for the purpose of model improvement in several applications experimentally, as well as discuss in detail how these effects are achieved.  We conclude the paper by providing {\bf practical recommendations} and discussing the opportunities and pitfalls in using explanations to improve deep learning models.

\section{From Explaining Predictions to Improving Models}
\label{sec:background}
This section offers an overview over existing explanation methods, and discusses how they can help improve machine learning models. Through various toy examples, this effect is showcased in an exemplary fashion, and finally formalized to build a basis for categorizing the various approaches presented in Section~\ref{sec:augmentation}.

\subsection{Global vs.\ Local XAI}
\label{sec:background:xai}
\glspl{dnn} are highly complex models with an enormous number of parameters that interact in a nonlinear fashion. While their learned representations perform potentially better than hand-crafted ones, the underlying decision-making cannot always be easily interpreted. 
To cope with this problem the field of explainable AI has recently developed a multitude of methods, which elucidate on and visualize a model's reasoning. Generally, two ends of the spectrum of \gls{xai} approaches can be distinguished:

\emph{Global} \gls{xai} focuses on interpreting a model's behavior and the features it has learned or is sensitive to in a general manner, \eg, by identifying and visualizing encoded concepts, as well as learned representations or sensitivities. Some methods aim to link information about specific, human-interpretable semantic concepts, \eg, labels, to single neurons \cite{Bau2017Network} and subnets \cite{Hu2020Architecture}, or to vectors in feature space \cite{Kim2018Interpretability}. In contrast to finding representations (in terms of neurons or filters) for pre-defined concepts, methods such as SUMMIT~\cite{Hohman2020Summit} instead generate neuron importance maps from which encoded concepts can be inferred, or hierarchical relationships between concepts visualized.

\emph{Local} \gls{xai} instead aims to explain the individual predictions of a model. That is, for specific samples, local explanation techniques seek to determine the parts of the input that have the most influence on a model's decision for one particular sample. Methods that modify the backward pass, such as \cite{Baehrens2010How, Bach2015Pixel, Montavon2017Explaining, Shrikumar2017Learning, Sundararajan2017Axiomatic}, explain efficiently in terms of computation time, but require grey-box access to a model's internal parameters. Sampling-based techniques leverage local surrogate models \cite{Ribeiro2016Why}, input perturbations \cite{Zintgraf2017Visualizing,Fong2017Interpretable} or even game theory \cite{vstrumbelj2014explaining,Lundberg2017Unified} to interpret complete black-box models. In turn, however, they consume far more computational resources, only arrive at estimated explanations (due to sampling), and do not offer intermediate attributions as part of the explanation process. For more information about the explanation methods, we refer the reader to the excellent review papers \cite{Samek2021Explaining, Tjoa2019Survey, Vilone2020Explainable}.

In the literature, explanation methods are frequently used for visualization purposes and to \textit{identify} problems with the model --- but not to alleviate them. Nevertheless, a variety of approaches following above goal exists, that we aim to review and systematize in this work. Most of these techniques focus exclusively on employing local \gls{xai}, due to the detailed and sample-specific nature of these explanations.

\subsection{Improving Different Model Properties with \gls{xai}}
\label{sec:background:improving}
When \gls{ml} models are deployed to real-world scenarios, they should not only achieve a good prediction accuracy, but also be trustworthy and reliable in their decision-making. However, traditional optimization metrics such as the test accuracy on a benchmark dataset do not guarantee that trustworthiness or reliability are fulfilled. As we will demonstrate on toy experiments (and also discuss theoretically in Section~\ref{sec:background:improving:theory}), explanations may help to improve on a variety of model properties, the most prominent of which we discuss in the following:

\begin{itemize}
    \item \textbf{Performance}:
    \gls{ml} models are usually trained in order to maximize their ability to predict correctly and with a high generalization ability. However, metrics solely focusing on the model's performance, such as test accuracy on benchmark datasets, may be difficult to maximize, as many tasks and datasets are extremely complex. %
    Existing models often have difficulties to thoroughly understand such datasets, and to identify the most descriptive --- and at the same time domain-relevant --- input features. Instead, detrimental effects such as overfitting on well correlating but comparatively domain-irrelevant input features in the training set may occur, and affect the test performance (and generalization beyond that) negatively. In this paper, we refer to test performance as \emph{Performance}. Note, however, that this property may not always reflect the model's actual generalization ability, \eg, if the test domain is too similar to the training domain (see \emph{Reasoning} property).
    \item \textbf{Convergence}: 
    Due to their enormous number of trainable parameters, modern \gls{ml} architectures usually take significant effort and time to train. Faster convergence is thus desirable, but often difficult to balance with converging to an optimum that achieves state-of-the-art performance.
    \item \textbf{Robustness}: 
    \glspl{dnn} are often overly sensitive to changes in the input --- even if those changes are imperceptible by humans. This effect can lead to adversarial cases where slight perturbations to input samples cause vastly different predictions \cite{Szegedy2014Intriguing, Papernot2016Limitations}, or make models untrustworthy, as the explanations for their decisions can be manipulated arbitrarily by perturbing inputs without affecting the prediction \cite{Dombrowski2019Explanations, Anders2020Fairwashing}. Both effects can be mitigated by increasing model robustness against slight alterations of the input. 
    \item \textbf{Efficiency}: 
    Current \gls{dnn}-architectures usually require enormous amounts of data to learn from in order to achieve state-of-the-art performance. Depending on the data domain, it either needs to be labeled manually by experts, incurring extremely high effort and cost, or it can be annotated using automation or crowdsourcing, with the danger of including wrong labels. Neither high labeling cost nor a model learning the wrong information can be considered efficient. Furthermore, a large number of parameters --- far more than are required in theory to solve the task at hand --- is beneficial during training, as it leads to alternative subnets, helps represent complex functions and makes models easier to optimize \cite{Simonyan2015Very, Krizhevsky2017Imagenet, Zhu2019Learning, Zou2020Gradient}. However, after a model is trained, a majority of these parameters does not effectively contribute much to solving the designated task accurately \cite{Yeom2019Pruning, Sabih2020Utilizing, Wang2021Emerging}, but nevertheless requires a lot of storage resources and increases time and energy cost during inference. Therefore, making models more efficient by reducing the amount of data or features required during training, as well as the number or storage cost of parameters after training --- while keeping performance intact --- is extremely desirable, as the increased efficiency eases many future applications, such as employing \glspl{dnn} on mobile devices. 
    \item \textbf{Reasoning.} 
    Due to their strongly data-dependent nature, modern \gls{ml} models are prone to reflect any regularities present in the training data --- whether these generalize to real-world circumstances or not. Therefore, even if (seemingly) performing well with a high test accuracy, the decisions of \gls{ml} models are often based on spurious correlations, biases, and \glsdesc{ch} features occurring in the training dataset \cite{Stock2018Convnets, Lapuschkin2019Unmasking, Anders2022Finding}. This is because all of these are features may appear helpful in solving a given task, without needing to understand the desired, potentially more complex (but often more valid) connections. Therefore, \emph{Reasoning} is connected to the \emph{Performance} property discussed above: If these confounding (or simply undesired) features are only present in the training set, such behavior can negatively affect test performance, and thus can be easy to identify. In this case, improving \emph{Reasoning} would simultaneously increase \emph{Performance}. On the other hand, if an undesired feature is present in both training and test sets, as is often the case, it may be helpful to the model and even increase test performance. The latter effect leads to test performance vastly overestimating a model's actual generalization ability and obscures the undesired model behavior. Improving \emph{Reasoning} in this case may impact \emph{Performance} negatively \cite{Anders2022Finding}, even though the resulting model is able to generalize better. As such, better \emph{Reasoning} is extremely difficult to measure reliably, and often depends on (human-defined) ground truths of feature desirability and importance.
    \item \textbf{Equality}:  
    Many real-world datasets contain inherent imbalances, \eg, between classes. When optimized to accurately predict such imbalanced datasets, machine learning models are prone to ignore the minority classes in favor of majority classes, since majority classes affect the average performance on the whole dataset the most and contribute significantly to the loss signal steering the training process. To achieve a good performance in the general case, equal treatment of all populations and sources of data has to be ensured. Note that this property is different from the notion of fairness, which is a widely researched topic on its own that we do not touch upon in this work.
\end{itemize}

Improvements \wrt\ the discussed model properties are crucial for the widespread and successful application of complex ML models. However, this is often far from trivial, due to the black-box nature of large models, and the resulting lack of information about the model and its decision-making. If exploited successfully, explanations can provide the additional information needed to improve above properties (at least partially). For instance, in the case of performance and convergence, knowledge about relevant and irrelevant feature representations can help focus on the most important ones and therefore reduce training time and increase accuracy. Similarly, identifying the most important neurons and filters is key to improving model efficiency. In order to gain an intuition of this phenomenon, we demonstrate empirically how \gls{xai} --- under the right conditions --- can help improve specific model properties through a series of simple toy experiments, with three different scenarios. Details about the experimental setup of the toy experiments can be found in \ref{sup:sec:toydetails}, and the datasets used in these experiments are visualized in \ref{sup:sec:toyfig}.

\subsubsection*{Toy Experiment 1 (Model Performance)}
\noindent The first experiment demonstrates that XAI can help improve prediction accuracy by XAI-based weighting of feature representations during the forward pass. More precisely, we use explanations to identify and increase intermediate features, which correspond to positive evidence for a specific prediction, and decrease the ones corresponding to negative evidence, similar to the approaches discussed in Section~\ref{sec:augmentation:feature}. In short, we trained a neural network to solve a binary classification task on a dataset with two informative and three random input dimensions (Dataset visualized in Figure \ref{fig:toy-dataset-1}).
We weighted intermediate features during the forward pass (similar to \cite{Sun2020Explanation}) using a feature-wise (e.g., intermediate input features to layer $l$) and sample-wise (e.g., obtained for sample $x_i$ with the network parameters $\theta^t$) attention mask $M_{\text{feat}}^{i, l, t} \in [0.5, 1.5]$. The weighted features are obtained as 
\begin{equation}
{f^l_{\theta^t}(x_i)}' = M_{\text{feat}}^{i, l, t} \odot f^l_{\theta^t}(x_i),
\end{equation}
with $\odot$ denoting the element-wise product. Thus, local explanations are used to construct $M_{\text{feat}}^{i, l, t}$, which acts as an attention filter to enhance or inhibit the relevant or irrelevant feature representations, respectively. In the experiment we compare the attention-augmented models with an unaugmented baseline. 

\begin{figure}[!h]
    \centering
    \includegraphics[width=0.95\linewidth]{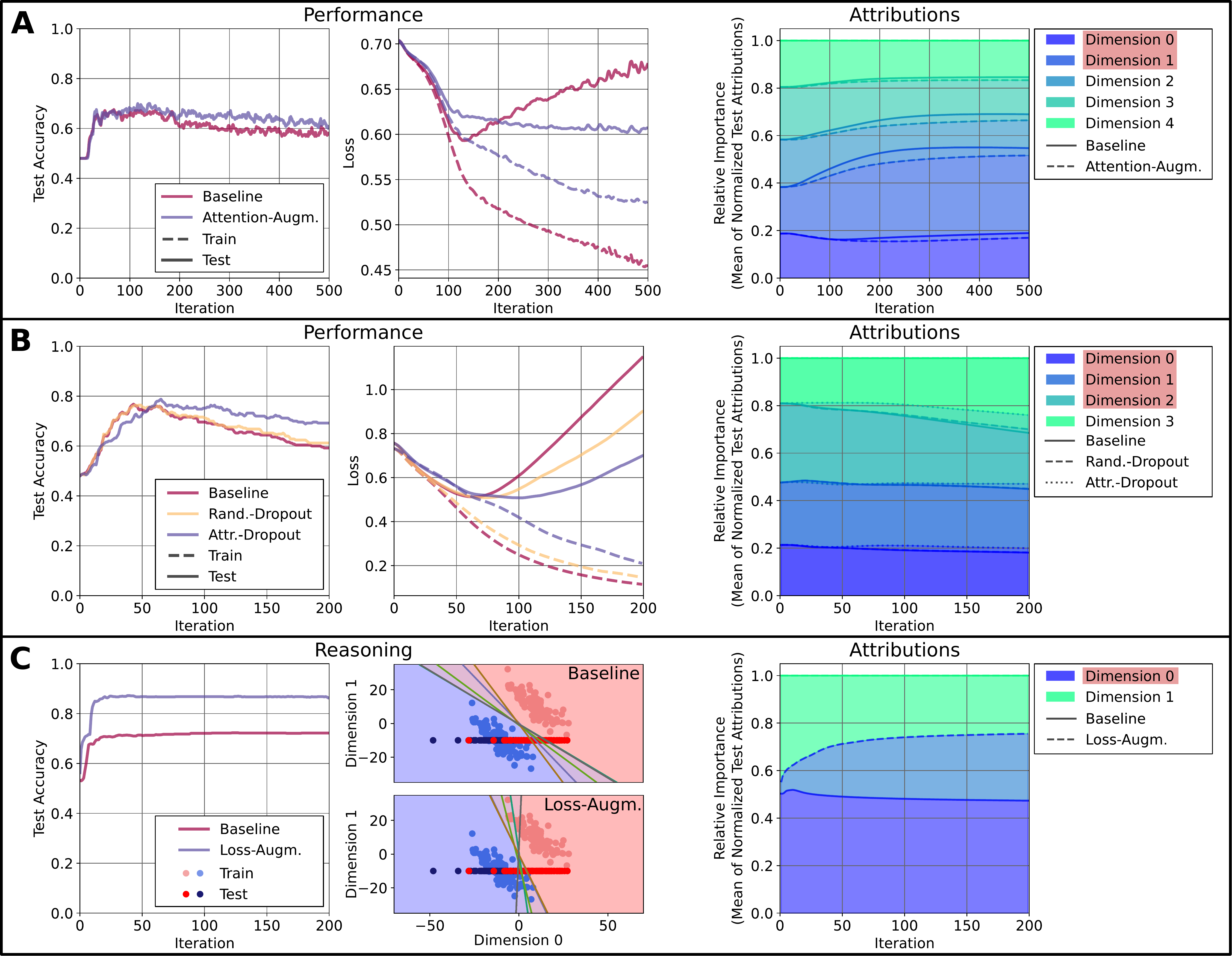}
    \caption{Toy experiments demonstrating the opportunities of \gls{xai} for model improvement. \textbf{A}: Improving performance by masking intermediate features during the forward pass (attention-augmentation) based on attributions, i.e., local explanations. The respective informative or desired input dimensions are highlighted in red. \emph{Left}: Attention-augmentation improves test accuracy over baseline. \emph{Center}: Attention-augmented models demonstrate a significantly reduced tendency to overfit. \emph{Right}: The augmented model puts a more equal relative importance on all input dimensions. \textbf{B}: Reducing overfitting by introducing explanation-guided dropout. \emph{Left}: Explanation-guided dropout improves test accuracy over baseline and random dropout models. \emph{Center}: Strong overfitting tendencies for the baseline models, slightly less for the random dropout models, but far less for the explanation-guided dropout models. \emph{Right}: Random dropout models attribute less importance to the distractor dimension 3 than the baseline. The explanation-guided dropout models exhibit this effect far more pronounced. \textbf{C}: Altering reasoning by using \gls{xai} to regularize the loss function. \emph{Left}:  The loss-augmented models significantly outperform the baseline models. \emph{Center}: Visualization of the train (pastel colored points) and test (saturated colored points) datasets, along with the achieved decision boundaries by the five repetitions of the experiment. The loss-augmented models barely use the undesired dimension 1 for their predictions. \emph{Right}: By regularizing the attributions in the loss function, the relative importance scores of dimension 1 are in fact reduced enormously.
    }
    \label{fig:toy-experiments}
\end{figure}

The results (based on 5 repetitions) of this experiment are depicted in Figure~\ref{fig:toy-experiments}A. The attention-augmented models are on average able to generalize considerably better than the baseline models that were trained without a feature re-weighting (see \emph{top left}). The trajectory of the test accuracy is also far more stable for the augmented models, and decreases less over training iterations. This finding is strongly supported by the associated training and test losses (\emph{center}). The baseline models achieve a lower training loss than the augmented models, but their test loss starts increasing from around iteration 120 --- a clear sign of overfitting. In contrast, the test loss of the augmented models stays comparatively constant from iteration 120 onward, even decreasing slightly but steadily. The normalized test set attribution values as shown on the \emph{right} explain the reduced overfitting of the augmented models. The relative importance of the three uninformative input dimensions (2--4) decreases steadily for both baseline and attention-augmented models. Between the two informative dimensions (0 and 1), the baseline models seem to strongly favor dimension 1. While this is still the case for the attention-augmented models, they attribute less importance to dimension 1 in comparison, using both informative dimensions more equally. However, the relative importance of dimension 0 also decreases compared to the baseline, while the relative importance of the uninformative dimensions increases. It seems that by quickly identifying the informative dimensions, and overfitting on them, the baselines simultaneously lose generalization performance. In contrast, the augmented models learn more cautiously but with increased stability, and thus are able to generalize better (note that considering all informative dimensions is a property of large margin classifiers).

\subsubsection*{Toy Experiment 2 (Model Performance)}
\noindent The second toy experiment also aims at improving model performance, this time by explicitly reducing overfitting on singular distractive input dimensions. We train a neural network to solve a binary classification task and generate the data in a way that the dimensions 0--2 are truly informative about the class, while dimension 3 indicates the correct class via the sign for the training samples, but is randomized for the test samples (Dataset visualized in Figure \ref{fig:toy-dataset-2}). A generalizing model would not only rely on the distractor dimension 3 for its predictions, but instead leverage information from all informative input dimensions (large margin property). In order to ensure this property, we use an explanation-guided dropout method, similar to \cite{Zunino2021Excitation}, which temporarily turns off the features that the model uses most to make its predictions. We compare this XAI-supported approach with the unaugmented baseline models and models that employ random (\ie, the standard) dropout.

Figure~\ref{fig:toy-experiments}B visualizes the results of this series of experiments. Due to how we designed our test set, the test performance shown to the \emph{left} is related to how much the three informative input dimensions, rather than the distractive fourth one are used. The random dropout improves slightly upon the baseline, indicating that the resulting models are less reliant on the fourth input dimension. While the explanation-guided dropout performs worse than both baseline and random dropout for the first few iterations --- as is to be expected, since learning would slow down when the currently important intermediate features are dropped out --- from around iteration 60, it outperforms even the random dropout increasingly. Concurrently with these observations, the training and test loss curves depicted in the \emph{center} indicate strong overfitting for the baseline models, since test losses start rising again around iteration 75. The same effect occurs for the random dropout and \gls{xai}-dropout, but far later (iteration 80 and 125, respectively), and less rapidly.
In order to gain an intuition about why above effects occur, we investigate the corresponding attributions in further detail. Specifically, the mean over samples and iterations of input dimension-wise test set attributions is shown to the \emph{right} of Figure~\ref{fig:toy-experiments}B. Compared to the baseline, the random dropout model attributes slighly less relevance to the distractor dimension relative to the informative dimensions. Again, this effect is even more pronounced for the explanation-guided dropout models.

Dropout is a useful tool in order to decrease overfitting, since it encourages the model to find alternative solutions to a given task \cite{Srivastava2014Dropout}. Usually, the dropped out neurons are chosen at random, which only stochastically encourages alternative solutions. In contrast, \gls{xai} allows for a smarter choice. By dropping out the most important neurons, the most relevant path through the network is always disrupted, maximizing the resulting increase in generalization. Note, however, that this effect may potentially be detrimental to learning on more complex tasks, since the model may never get a chance to converge properly. Making the dropout criterion increasingly random over the course of training may therefore be beneficial.

\subsubsection*{Toy Experiment 3 (Model Reasoning)}
\label{sec:background:improving:toy3}
\noindent For the last toy experiment, we aim at aligning model reasoning to a ground truth provided by a human expert, similar to the approaches discussed in Section~\ref{sec:augmentation:loss}. We train models on two-dimensional data, and alter their decision-making to ignore one of the two dimensions. This selection seems arbitrary in a toy setting, however, even on complex datasets, the decision whether an input feature is desired or not often requires a human expert rather than being made automatically based on the available data alone. The data is visualized in Figure~\ref{fig:toy-experiments}C (\emph{center}), with the training set consisting of the pastel colored points and the test set in saturated colors, and in Figure \ref{fig:toy-dataset-3}. While the training set varies in the direction of dimension 1, the test set does not. After computing the explanations, we augment the loss function similarly to \cite{Ross2017Right} as 

\begin{align}
& \mathcal{L}_{\text{loss-aug}}^{l, t}(x_i) = \mathcal{L}_{\text{pred}}(f_{\theta^t}(x_i), y_i) + \mathcal{L}_{\text{reason}}(r_i, r_A), \\
\text{with }\ & \mathcal{L}_{\text{reason}}(r_i, r_A) = ||(1 - r_A) \odot {(r_i')^{l, t}}||_2^2,
\label{eq:toy-aug:loss}
\end{align}
where $\mathcal{L}_{\text{pred}}$ is the standard classification loss, and $r_A$ a binary ground truth mask (same one for the whole dataset). Through the regularization term, the model is rewarded for aligning its explanations ($(r'_i)^{l,t}$ before layer $l$ at training iteration $t$) with the ground truth explanations ($r_A$). More precisely, it is penalized for using input dimensions marked as not desirable (i.e., $r_A = 0$).

The results of this experiment are visualized in Figure~\ref{fig:toy-experiments}C. Due to the design of the test set, which only varies in the direction of dimension 0, the test accuracy over iterations depicted on the \emph{left} directly indicates how well a model can classify without relying on the undesired dimension 1. Here, the loss-augmented models quickly and consistently outperform the baseline models, implying that the loss-augmented models rely less on input dimension 1. This interpretation is confirmed by the corresponding decision boundaries (\emph{center}): The boundaries of the loss-augmented models are shifted to be nearly completely orthogonal to dimension 0, demonstrating that the models' decision-making is changed to exhibit the desired behavior. While the baseline models attribute almost equal importances to both input dimensions (\emph{right}), the loss regularization leads to a significant increase for the importance of dimension 0,  while the opposite is the case for the undesired dimension 1.

In this experiment, \gls{xai} provides a measurement for model behavior, which can be compared against a ground truth, and in turn employed to alter model behavior as desired. But for more complex tasks, even if the explanations change through loss regularization, there is no guarantee that the decision behavior is altered in the desired manner, as indicated by \cite{Dombrowski2019Explanations, Anders2020Fairwashing}. Due to the simple setting, however, this effect cannot occur in the above toy experiment. Ground truth explanation(s) $r_A$ are required for adapting reasoning through loss regularization, but obtaining them requires involvement from human experts and may be infeasible for large datasets and complex tasks in practice.\\

As demonstrated through the three example scenarios, \gls{xai} can be employed in various ways to improve different model properties, since it is able to measure and identify the relevance of input dimensions, features, and neurons wrt.\ the model's decision-making. Therefore, it is not only useful for informing human experts about a model's behavior, but can be utilized practically to obtain better models.

\subsection{Theoretical Formalization}
\label{sec:background:improving:theory}
After showcasing the ability of \gls{xai} to improve \gls{ml}-models empirically through different toy examples, we will formalize those insights in order to derive a theoretical framework for categorizing existing approaches in a systematic manner.

\begin{figure}
    \centering
    \includegraphics[width=\linewidth]{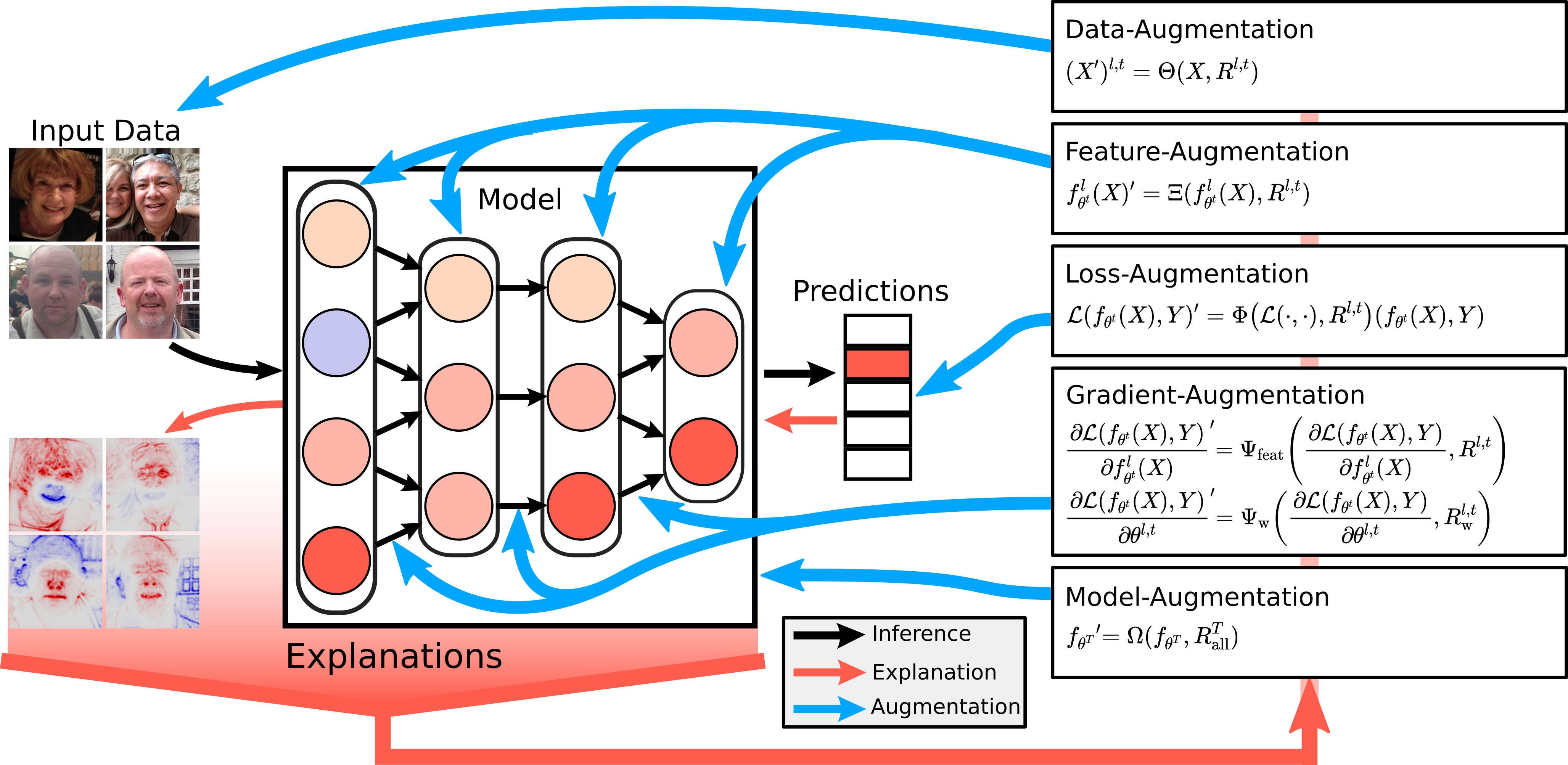}
    \caption{Model improvement with \gls{xai}. Explanations offer information about the model decision-making and behavior, which may in turn be leveraged to improve various model properties by augmenting different components of the training process or by adapting the trained model.}
    \label{fig:overview}
\end{figure}

Assume we have a model $f_{\theta^t}$, parametrized by parameters $\theta^t$ after training iteration $t \in  \{1, ..., T\}$, and some data $X = \{x_1, ..., x_N\}$\footnote{Note that we switch to dataset-wise notation in this Section, since it allows for more concise and readable expressions, as opposed to the sample-wise notation in other Section that is required due to details of some augmentation methods being sample-specific.} with ground-truth labels $Y = \{y_1, ..., y_N\}$. The model $f_{\theta^t}$ consists of $L$ layers $l \in \{0, \cdots, L-1\}$, where the parameters of each singular layer after $t$ training iterations are denoted by $\theta^{l, t}$. The \emph{input features} to layer $l$ are given by $f^l_{\theta^t}(X)$, so that, \eg, the input can be written as $X = f^0_{\theta^t}(X)$, and the model's output as $f_{\theta^t}(X) = f^{L}_{\theta^t}(X)$.
We further assume that a (local) \gls{xai} technique is available, that is able to provide explanations $R^{l, t} = \{r_1^{l, t}, ..., r_N^{l, t}\}$ for the model's decisions at each intermediate layer $l$ and iteration $t$, corresponding to intermediate features $f^l_{\theta^t}(X)$. These explanations can be leveraged to augment each component --- \ie, Data $\rightarrow$ Feature Representations $\rightarrow$ Loss Function $\rightarrow$ Gradient $\rightarrow$ Trained Model --- separately. Some examples of this approach were discussed in the toy experiments in Section~\ref{sec:background:improving}. The different types of augmentation at each component of the training loop are depicted in Figure~\ref{fig:overview}.\\

\begin{figure}[ht]
    \centering
    \includegraphics[width=0.85\linewidth]{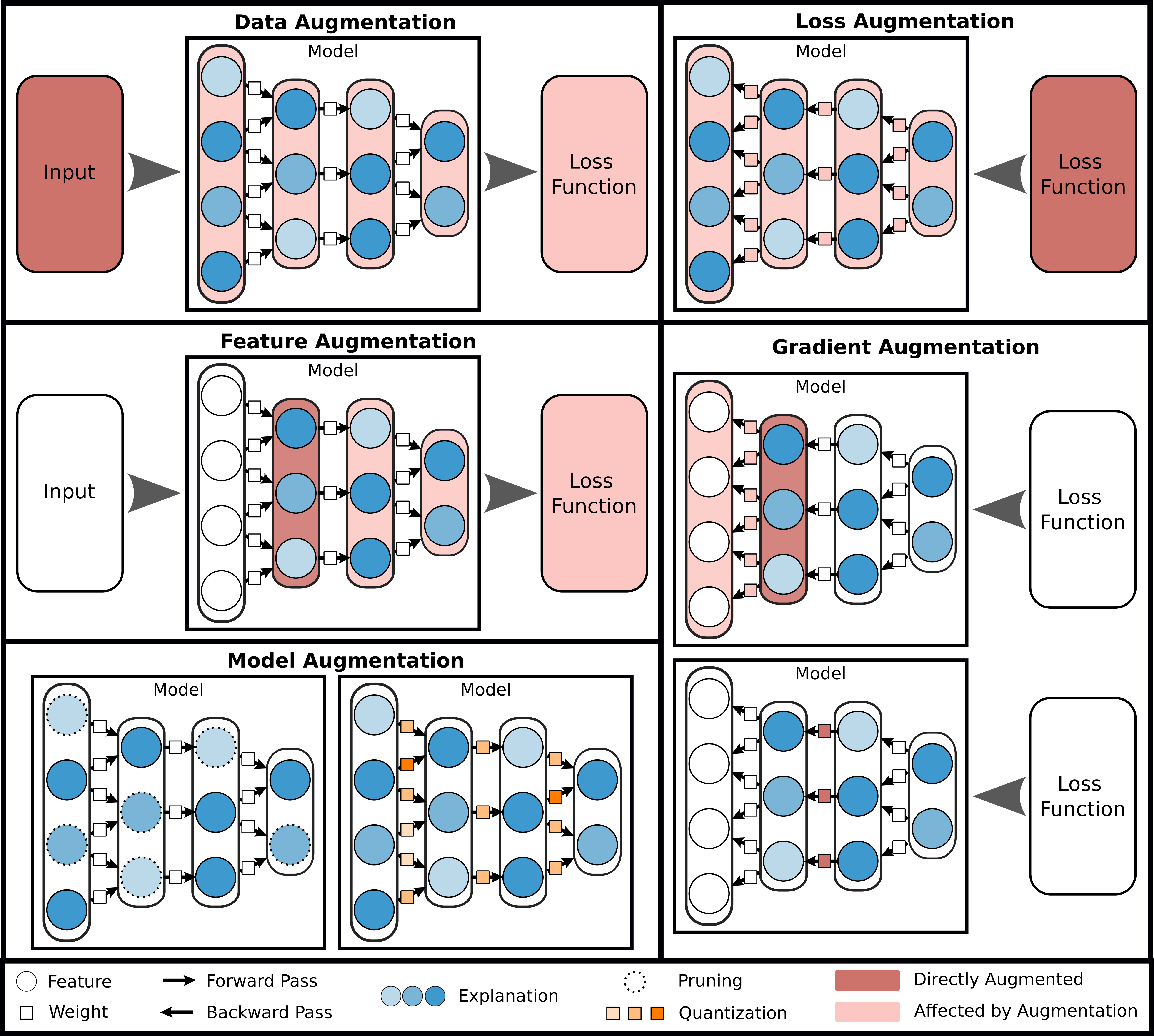}
    \caption{Types of \gls{xai}-based augmentation. \emph{Top left}: Data augmentation alters the input distribution by utilizing input layer explanations. It therefore affects all components of the training loop as well as the final model. \emph{Center left}: Feature augmentation leverages intermediate explanations at a specific layer to mask or transform the corresponding intermediate features, indirectly affecting all higher feature representations and subsequent components of training, such as gradient updates, and the resulting model. \emph{Top right}: Loss augmentation uses the performance measuring capabilities of the loss function and affects all components of the backward pass and the final model indirectly. \emph{Bottom right}: Two sub-types of gradient augmentation exist: Methods that augment the feature gradient at a specific layer (\emph{top}) by masking or transforming it affect all parameter updates and feature gradients of lower layers in addition to the final model. In contrast, approaches that augment the parameter gradient at a specific layer (\emph{bottom}) instead only affect the parameter updates of this layer. However, parameter-wise explanations are required for this purpose, which not all \gls{xai} techniques are able to easily provide. \emph{Bottom left}: Augmentation of the model after training. Explanations at intermediate layers are leveraged to estimate neuron, filter, or parameter importance, and employ it as a criterion for, \eg, pruning (\emph{left}) or quantization (\emph{right}).}
    \label{fig:augmentation_types}
\end{figure}

\noindent\textbf{Augmenting the Data.}
\gls{xai}-dependent data augmentation leverages explanations in order to alter the structure of the data. We describe this alteration via the general function $\Theta(X, R^{l, t})$, which takes the original data $X$, and corresponding attributions $R^{l, t}$ as inputs, in order to generate augmented data as
\begin{equation}
(X')^{l, t} = \Theta(X, R^{l, t})~.
\label{eq:augmentation:data:base}
\end{equation}
In practice, $\Theta$ either generates new samples depending on $R^{l, t}$ \cite{Teso2019Explanatory, Schramowski2020Making}, or alters the distribution of existing data \cite{Weber2020Towards}.
As shown in Figure~\ref{fig:augmentation_types} (\emph{top left}), this type of augmentation is applied during the first component of the forward-backward training loop, and therefore affects all other components.\\
        
\noindent\textbf{Augmenting the Intermediate Features.}
As explanations can provide a measurement of feature importance, this information can be leveraged to scale, mask, or transform intermediate features, as shown by the toy experiments in Section~\ref{sec:background:improving} and the approaches discussed in Section~\ref{sec:augmentation:feature}. We generalize these augmentations using the function $\Xi(f^l_{\theta^t}(X), R^{l, t})$, which takes the input features $f^l_{\theta^t}(X)$ to layer $l$ and the corresponding attributions $R^{l, t}$ as inputs, in order to generate augmented feature representations as 
\begin{equation}
f^l_{\theta^t}(X)' = \Xi(f^l_{\theta^t}(X), R^{l, t})~.
\label{eq:augmentation:feature:base}
\end{equation}
As shown in Figure~\ref{fig:augmentation_types} (\emph{center left}), augmenting intermediate features does not affect the model's inputs, but all subsequent training process components. 
Note that the input space can technically be viewed as an extension of the feature space. However, as will be shown in further detail in Section~\ref{sec:augmentation:data}, the relevant data augmentation approaches focus on altering the distribution of the data \emph{as a whole} --- as opposed to most (intermediate) feature augmentation approaches, that focus on masking or transforming feature representations on a \emph{per-sample} basis.\\
        
\noindent\textbf{Augmenting the Loss Function.}
The loss function determines the behavior of a model. Thus, augmenting the loss function based on explanations can help specify which behavior is desired, using explanations as a feedback. Based on the standard loss function $\mathcal{L}(f_{\theta^t}(X), Y)$, measuring a model's predictive error, and explanations $R^l$, the general augmented loss function is then given as the output of augmentation $\Phi$, \ie
\begin{equation}
\mathcal{L}(f_{\theta^t}(X), Y)' = \Phi \left( \mathcal{L}(\cdot, \cdot), R^{l, t} \right) (f_{\theta^t}(X), Y).
\label{eq:augmentation:loss:base}
\end{equation}
For instance, $\Phi$ can either add an \gls{xai}-based regularization term or scale class-wise losses, as discussed in Section~\ref{sec:augmentation:loss}. Augmenting the loss function in this manner only affects the backward pass (see Figure~\ref{fig:augmentation_types} (\emph{top right})).\\

\noindent\textbf{Augmenting the Gradients.}
Similarly to how explanations can be leveraged in order to augment feature representations during the forward pass, the information about feature importance that they offer is applicable to the backward pass as well. Here, however, two sub-types of augmentation can be distinguished, due to how parameter updates are derived using the chain rule.
    
Firstly, as depicted at the \emph{top} of the \emph{bottom right} panel of Figure~\ref{fig:augmentation_types}, the intermediate feature gradients at layer $l$ can be scaled, masked, or transformed, mirroring the previously discussed feature augmentations during the forward pass. Based on the corresponding attributions $R^{l, t}$, we formulate this augmentation generally as 
\begin{equation}
\frac{\partial \mathcal{L}(f_{\theta^t}(X), Y)}{\partial f^l_{\theta^t}(X)}' = \Psi_{\text{feat}}\left(\frac{\partial \mathcal{L}(f_{\theta^t}(X), Y)}{\partial f^l_{\theta^t}(X)}, R^{l, t}\right)~.
\label{eq:augmentation:feature-gradient:base}
\end{equation}
When applied to an intermediate layer $l$, this technique additionally alters feature gradients and parameter updates of all layers lower than $l$.
    
Alternatively, the parameter gradients can be augmented directly by computing parameter-wise importance scores $R_{\text{w}}^l$. These can either be obtained from the intermediate feature explanations $R^{l, t}$ and $R^{l+1, t}$ \cite{Lee2019Improvement}, or, in the case of many modified backpropagation \gls{xai} approaches, such as \gls{lrp} \cite{Bach2015Pixel}, by simply computing explanations \wrt\ the parameters \cite{Becking2021Ecq}. The corresponding augmentation can then be generalized as
\begin{equation}
\frac{\partial \mathcal{L}(f_{\theta^t}(X), Y)}{\partial \theta^{l, t}}' = \Psi_{\text{w}}\left(\frac{\partial \mathcal{L}(f_{\theta^t}(X), Y)}{\partial \theta^{l, t}}, R_{\text{w}}^{l, t}\right)~.
\label{eq:augmentation:parameter-gradient:base}
\end{equation}
In contrast to Equation~\eqref{eq:augmentation:feature-gradient:base}, this only leads to the parameter updates at layer $l$ being altered, as visualized at the \emph{bottom} of the \emph{bottom right} panel of Figure~\ref{fig:augmentation_types}. Approaches that augment the gradients during the backward pass are further discussed in Section~\ref{sec:augmentation:gradient}.\\

\noindent\textbf{Augmenting the Model.}
Even after a model is trained, the information \wrt\ intermediate feature importance offered by \gls{xai} may still be leveraged in order to augment the whole model, \eg, in order to alter its structure or reduce the amount of storage space required by the parameters. More concisely, using explanations $R_{\text{all}}^T = \{R^{0, T}, ..., R^{L-1, T}\}$ before each layer of the model and after the last iteration $T$, the model $f_{\theta^T}$ can generally be augmented as
\begin{equation}
f_{\theta^T}{}' = \Omega(f_{\theta^T}, R_{\text{all}}^T)~.
\label{eq:augmentation:model:base}
\end{equation}
In practice, \gls{xai} is either employed here to prune (Figure~\ref{fig:augmentation_types}, \emph{left} of the \emph{bottom left} panel) or quantize the model (Figure~\ref{fig:augmentation_types}, \emph{right} of the \emph{bottom left} panel), as discussed further in Section~\ref{sec:augmentation:model}.
Note that typically in literature, the above categories of \gls{xai}-based augmentation are only applied one at a time. However, due to each category altering a separate component of the training process, in theory, multiple augmentations, \eg, targeting the same model property, could be applied at the same time, altering different components of the training process simultaneously.
    
\section{Review of Methods for XAI-Based Model Improvement}
\label{sec:augmentation} 
This section reviews different types of methods proposed for XAI-based model improvement. A main conceptual difference, which allows to distinguish these methods, is the component of the training loop at which augmentations are performed in order to achieve the improvement. More precisely, changes can be applied at the dataset-level, during the forward pass to the feature representations, within the loss function, to the gradients during optimization, or after training to the model structure and parameters, as discussed in the previous section. Generally, each of these augmentations comes with different implications in terms of the improved model properties, computational costs, time requirements, and model agnosticity. 

\subsection{Augmenting the Data}
\label{sec:augmentation:data}
The fitting and training processes in \gls{ml} are generally highly data-dependent, especially for the widely employed \glspl{dnn}, with the performance of resulting models being strongly affected by the available input features, imbalances, biases, and spurious correlations present within the dataset. Additionally, (local) \gls{xai} methods primarily (although not exclusively) focus on providing explanations in input space, as they target a human audience. When employing \gls{xai} to improve \gls{ml} models, augmentations of the data as described in \eqref{eq:augmentation:data:base} thus come naturally to mind. In principle, methods not discussed in this Section, particularly some feature augmentation methods (Section~\ref{sec:augmentation:feature}) could be applied to the input space as well (as is done, for instance, in \cite{Anders2022Finding}). However, here we distinguish data augmentation approaches as those that focus on changing the distribution of the data as a whole --- as opposed to altering singular samples.

By changing the distribution of samples that are fed to the model, the approaches in this category aim to mitigate biased, unfair, or simply wrong decision-making behavior that is caused by the vanilla data structure. For this purpose, explanations can be leveraged in order to generate artificial samples that provide exemplary information against undesired behavior \cite{Teso2019Explanatory, Schramowski2020Making}, or to simply re-sample the existing data to achieve fairer predictions. 
More precisely, \gls{xil}~\cite{Teso2019Explanatory, Schramowski2020Making} enables human users to correct a model's decision-making instance-wise during an active learning setting. Here, the learner is able to periodically query a user for the correction of its prediction. In order to improve model reasoning and human trust in a model's decisions, \gls{xil} leverages local explanations: After selecting an informative instance, the subsequent query from model to user here includes the instance, corresponding prediction, as well as an explanation for this decision in form of a heatmap. In turn, the user may not only correct the prediction, but the explanation as well, and provide this feedback to the model. Data augmentation is then performed based on the corrected explanation by generating counter examples to discourage the model from depending on irrelevant input features. \gls{xil} thereby corrects the model's reasoning, avoids \gls{ch} \cite{Goodfellow2016Cleverhans,Lapuschkin2019Unmasking} moments, and increases user's trust into the model's decision-making. The work in \cite{Schramowski2020Making} further extends \gls{xil} to alternatively augment the loss function using the \gls{rrr} loss proposed by \cite{Ross2017Right}, which will be further discussed in Section~\ref{sec:augmentation:loss}. The authors of \cite{Stammer2021Right} employ this loss-augmenting variant of \gls{xil} to Neuro-Symbolic concept learners, allowing for concept-based, semantic corrections to be made to the model.

The authors of \cite{Gautam2021This} also aim at improving model reasoning, specifically \wrt\ artifacts such as \gls{ch} effects or Backdoors \cite{Chen2019Detecting}, by discovering and removing artifactual samples from the dataset. To obtain explanations,  they employ the self-interpretable ProtoPNet \cite{Chen2019This} architecture, and extend this method to \gls{prp} using concepts from \gls{lrp} \cite{Bach2015Pixel}. Multiple explanations for each input image \wrt\ its true class can thus be obtained. Various multi-view clustering approaches, such as \cite{Kumar2011Coregularized, Zhan2019Multiview}, are subsequently employed to detect artifact samples and separate them from clean samples effectively. By training a new model on the clean data, the effects of specific artifacts on the model's decision-making can be mitigated. 

In order to improve model performance on visual data, specifically on fine-grained classification tasks (\ie, tasks with classes that have similar features, such as distinguishing different dog breeds, and thus require attention to detail), the authors of \cite{Bargal2019Guided, Bargal2021Guided} propose Guided Zoom. This method aims at refining model predictions through a comparison with evidence used at training time that led to correct decisions. First, an evidence pool is generated from the training data, by selecting the top-l salient patches of samples correctly classified by the original model $f$, based on an evidence grounding method. Specifically, explanations, \ie, contrastive Excitation Backprop \cite{Zhang2018Excitation} are employed as a grounding method here. An evidence model $f^e$ is then trained on the obtained evidence pool to solve the same task as $f$. During test time, for each test sample $x$, the original model's decision is refined by selecting the top-l salient regions \wrt\ the top-k predicted classes according to $f(x)$, and taking the prediction of $f^e$ on these regions into account through a weighted combination. Predictions are therefore refined by an additional comparison of evidence used by $f$ for a preliminary prediction to known class-specific evidence encoded via $f^e$. The authors of \cite{Bargal2019Guided, Bargal2021Guided} show that Guided Zoom improves model performance, and, potentially, model reasoning, as the refined decisions are less based on biases --- although confounders that are strong enough to be selected as positive evidence towards a specific class might still affect the final decisions.

In the context of imbalanced data, \cite{Weber2020Towards} introduce \gls{xai}-guided imbalance mitigation, aiming towards an equal classification performance between classes. For this purpose, scalar metrics are first derived from a few representative attribution maps (here \gls{lrp}) in order to extract descriptive information, such as the attribution map entropy, and the pairwise distance between attribution maps for the same sample at different times during training. These are observed to correlate strongly to test accuracy and F1-Score, and thus demonstrate that attributions allow for an estimation of a model's generalization performance and convergence, even if there is no labeled data available, and can distinguish well between accurately and inaccurately predicted classes. This insight is then exploited to achieve more balanced class-wise performances when training a model on imbalanced datasets, by augmenting the learning process based on representative attributions. In a second step, class-wise factors are computed from above metrics repeatedly during training. These are able to react to the model's decision-making in order to alter the distribution of the input data that is fed to the model, leading to faster convergence and more equalized class-wise performances. Similarly, class-wise factors are also used  to augment the loss by scaling its class-wise contributions, which will further be discussed in Section~\ref{sec:augmentation:loss}.

In the original works, \gls{lime} \cite{Ribeiro2016Why} is used as means to obtain \gls{xai} attributions for \gls{xil}, although the framework is formulated in a general manner and considers any technique that offers local explanations. Similarly, \cite{Weber2020Towards} employ \gls{lrp} to obtain explanations. Other (local) explanation methods would be applicable as well, as demonstrated in an exemplary fashion in Section~\ref{sec:demoexamples}. The authors of \cite{Bargal2019Guided, Bargal2021Guided} also formulate their method generally enough to account for explanation techniques beyond the employed Excitation Backprop. In contrast, the technique proposed by \cite{Gautam2021This} relies on \gls{prp}, a specific explanation method which makes a model self-explainable, instead of providing post-hoc attributions, but in turn requires the model to be altered with a specialized additional layer and subsequent adaptation of parameters. While \gls{xil} allows for extremely thorough and instance-wise corrections to be made to the model, the required human feedback, especially the explanation corrections, in turn make it not only extremely slow in practice, but also subject to human error. The technique of \cite{Weber2020Towards} increases the computational load during model training, due to the repeated calculation of attributions, yet the additional (real) time cost is negligible, since no human interaction is required. The technique of \cite{Gautam2021This} uses aggregated explanations and clustering algorithms to automatically separate artifactual and clean samples, drastically limiting the amount of required human involvement. However, they propose to remove the affected samples from the dataset, thus eliminating potentially valuable information in addition to the undesired artifacts. Guided Zoom requires no human involvement, but requires the training of an additional evidence model. Furthermore, inference time drastically increases due to each prediction requiring multiple inference steps, one for the original image and each selected salient region.

Due to the data-dependent nature of \gls{ml} models, the impact of data augmentation methods on the model --- and therefore their improvement potential --- is extremely promising. As they only require access to the input-space, approaches that augment the data are generally able to retain a relatively high degree of model agnosticity.

\subsection{Augmenting the Intermediate Features}
\label{sec:augmentation:feature}
While a model's parameterization directly depends on the given (training) data, the feature space is often able to encode this data in a more concise form, so that some problems, \eg~, with biases or reasoning, may be easier and more effectively (compared to input space) mitigated there via augmentation, as described in Equation~\eqref{eq:augmentation:feature:base}. Two different types of methods exist in this category.\\

\noindent{\bf Attention and Intermediate Feature Masking.} 
This type of \gls{xai}-guided feature augmentation techniques seek to boost a model's performance by using explanations to distinguish relevant intermediate features from irrelevant ones. For this purpose, \gls{xai} methods that can provide intermediate explanations are required. As the shape of these intermediate explanations matches the shape of the features they explain, they can be directly employed to obtain a mask that expresses feature importance and weights them during the forward pass, similar to an attention mechanism. As a consequence, the features of the augmented layer, as well as the features of all higher layers are affected. 
In the context of image recognition, \gls{abn} \cite{Fukui2019Attention} interprets the local explanations obtained by extending \gls{cam} \cite{Zhou2016Learning} as an attention map. \gls{abn} is comprised of three modules, a feature extractor, an attention branch, and a perception branch. While the perception branch is a standard classifier on the feature extractor's output, the attention branch uses the output of the feature extractor to compute an attention map. In its original formulation, \gls{cam} uses a triple of convolutional layer, \gls{gap}, and a fully-connected classification layer to return pixel-wise importance scores. These are obtained as the average activations over the channels of the convolutional layer, weighted by the impact of each channel on the classification score of a target class. Since this requires a trained model, the attention branch of \gls{abn} omits the fully-connected classification layer, and instead uses a convolutional layer with $K$ channels that each correspond to a category. It can thus learn the attention map by interpreting the \gls{gap} result for each convolutional channel (\ie, one channel per class) as a logit and applying softmax to obtain class-wise probability scores. Utilizing this attention map, the perception branch input (\ie, the input of layer $l$) can then be masked in order for the model to focus on the most important parts of a given sample. Since both branches output separate probability scores, the resulting \gls{abn} is trained using a loss function $\mathcal{L}_{\text{abn}}$ that simply sums up the branch-wise losses:
\begin{equation}
\mathcal{L}_{\text{abn}}(f_{\theta^t}(x_i), y_i) = \mathcal{L}_{\text{att}}(f_{\theta^t}(x_i), y_i) + \mathcal{L}_{\text{per}}(f_{\theta^t}(x_i), y_i)~,
\label{eq:abn:original_loss}
\end{equation}
where $\mathcal{L}_{\text{att}}$ and $\mathcal{L}_{\text{per}}$ are the attention and perception branch losses, respectively. With this method, \gls{abn} is able to simultaneously explain decisions and exploit this knowledge for improved performance. 

The authors of \cite{Mitsuhara2019Embedding} extend \gls{abn} by additionally correcting attention maps by exploiting human knowledge, and thereby improve upon the visual explanation and classification performance of the original approach, while also altering model reasoning. For this purpose, an \gls{abn} is first trained, and the attention maps corresponding to each sample are stored. A human expert then edits and corrects these attention maps, and the model is finetuned using the edited attention maps. In the last step, \emph{loss augmentation} (see Section~\ref{sec:augmentation:loss}) is performed additionally, regularizing Equation~\eqref{eq:abn:original_loss} to produce attention maps that are similar to the corrected ones:
\begin{align}
& \mathcal{L}_{\text{abn}}(f_{\theta^t}(x_i), y_i) = \mathcal{L}_{\text{att}}(f_{\theta^t}(x_i), y_i) + \mathcal{L}_{\text{per}}(f_{\theta^t}(x_i), y_i) + 
\gamma \mathcal{L}_{\text{reason}}(r_i, a_i), \\
\text{with}~ & \mathcal{L}_{\text{reason}}(r_i, a_i) = ||a_i^{l, t}-r_i^{l, t}||_2~,
\label{eq:abn:corrected_loss}
\end{align}
where $r_i^{l, t}$ is the \gls{abn} attention map,  $a_i^{l, t}$ the corrected attention map, $|| \cdot ||_2$ the $\ell_2$-norm, and $\gamma$ a scale factor.

Similar to \gls{abn}, \cite{Schiller2019Relevance} also aim at improving the image classification performance --- specifically whale sound spectrogram classification --- of \gls{cnn}-type architectures by masking the feature extractor output (and classifier input) based on local \gls{xai}. More precisely, \cite{Schiller2019Relevance} first feeds the whole dataset to a \gls{cnn} pre-trained on orca sound spectrograms. The intermediate activations after the feature extractor are then explained using the \gls{dtd} method \cite{Montavon2017Explaining}. From these attributions, binary masks are then computed with two different goals, setting either the least relevant or the most relevant feature representations to zero. New classifiers are trained on the masked features, or on varying concatenations of masked features across different models and masking methods, resulting in a significantly increased classification performance compared to the base models, especially for the concatenation of both masked feature representations. Furthermore, by reducing the amount of used features --- the authors employ a \emph{binary} mask which turns off a subset of features, in contrast to other approaches --- efficiency is increased while preserving performance.

The authors of \cite{Sun2020Explanation} apply a similar method in a few-shot setting. In few-shot classification, the goal is for a (pre-trained) model to generalize to new classes, using only a small number of examples. However, this generalization can be difficult, if the source and target domains are very different. To mitigate that problem, the explanation-guided training proposed by \cite{Sun2020Explanation} leverages \gls{xai}, more specifically \gls{lrp}: Assuming a model that can be split into feature processing and classifier, during a forward pass, the feature processor outputs feature representations $f^l_{\theta^t}(x_i)$ for each sample $x_i$, which are then used by the classifier to arrive at prediction $f_{\theta^t}(x_i)$. Using \gls{lrp}, $f_{\theta^t}(x_i)$ is explained in terms of the feature processing output $f^l_{\theta^t}(x_i)$, as $r_i^{l, t}$. Similarly to the approaches of \cite{Fukui2019Attention} and \cite{Schiller2019Relevance}, $r_i^{l, t}$ can the be employed --- after normalizing it to the interval $[-1, 1]$ to obtain $(r')_i^{l, t}$ --- as a mask, weighing the intermediate features  $f^l_{\theta^t}(x_i)$ as 
\begin{equation}
f^l_{\theta^t}(x_i)' = (1+(r')_i^{l, t}) \odot f^l_{\theta^t}(x_i),
\label{eq:few-shot:fplrp}
\end{equation}
with $\odot$ denoting the element-wise product. When feeding $f^l_{\theta^t}(x_i)'$ into the classifier, the \gls{lrp}-weighted prediction $f_{\theta^t}(x_i)'$ is obtained. The model can then be trained using an objective function that considers both predictions, \ie, 
\begin{equation}
\mathcal{L}(f_{\theta^t}(x_i), f_{\theta^t}(x_i)', y_i) = \alpha \mathcal{L}(f_{\theta^t}(x_i), y_i) + \beta \mathcal{L}(f_{\theta^t}(x_i)', y_i),
\label{eq:few-shot:objective}
\end{equation}
where $\alpha, \beta$ are positive scalars. By applying above method, \cite{Sun2020Explanation} are able to significantly improve model performance in cross-domain few-shot classification.

Similarly, \cite{Zunino2021Explainable} improve domain generalization performance by using \gls{xai}-feedback (specifically, \gls{gcam} \cite{Selvaraju2016Gradcam} is employed during their experiments) to force the model to exhibit the correct reasoning and focus on the correct objects, rather than contextual information. For this purpose, they require binary ground-truth annotation masks for each sample that localize these objects, which is rescaled accordingly if explanations are computed at an intermediate layer. Periodically during training, each training sample is explained, and the explanation is compared to the ground-truth annotation. For each sample, if the peak, \ie, the maximum value of the corresponding explanation lies within the object area marked by the annotation, a binary mask $M^{l, t}_i$ is defined, which is set to 1 where the explanation is larger than zero, and to 0 otherwise. Otherwise $M^{l, t}_i$ is equal to the (potentially rescaled) ground truth annotation. $M^{l, t}_i$ is then employed to select intermediate features as 
\begin{equation}
f^l_{\theta^t}(x_i)' = M^{l, t}_i \odot f^l_{\theta^t}(x_i).
\label{eq:zuninoexplainable}
\end{equation}

Intuitively, the approach of \cite{Zunino2021Explainable} can be understood as ``dropping in'' features that are seemingly correctly used by the model. In contrast, dropout aims at avoiding overfitting by periodically setting a subset of features to zero during training. For this purpose, random dropout \cite{Srivastava2014Dropout} is usually employed, which simply decides randomly which features to mask at a given iteration. The authors of \cite{Zunino2021Excitation} improve upon this technique by instead employing \gls{xai}, more specifically Excitation Backpropagation \cite{Zhang2018Excitation}, to drop out more important neurons with a higher probability in order to speed up the overfitting avoidance effect. During training, each sample is explained and a sample-specific dropout mask is computed accordingly, leading to not only an improved generalization ability, but also less degradation when neurons are removed, compared to random dropout.

All of the above methods --- with the exception of \cite{Mitsuhara2019Embedding} --- are applicable in a relatively efficient manner and without human interaction during learning, so that no additional time investment is required on top of the standard model training time --- although this time may be slightly increased due to the need for computing explanations for each sample during training, which is nevertheless comparatively insignificant when using, \eg, modified backpropagation approaches to explain. In contrast, \cite{Zunino2021Explainable} include human knowledge through the necessary ground-truth object segmentation masks for each sample, which can be extremely tedious and infeasible to obtain. The approach of \cite{Mitsuhara2019Embedding} instead requires manual editing of a multitude of attention maps by a human expert, which does not scale well to large datasets, but improves upon the original approach of \cite{Fukui2019Attention} in terms of explanation, reasoning, and performance. The inclusion of human knowledge by \cite{Mitsuhara2019Embedding} further mitigates the issue that the attention may be inconsistent with the ground truth, \eg, by including multiple objects, since \gls{abn} learns the attention map instead of computing it \wrt\ a target class from a trained model as in the original formulation of \gls{cam}. While the model structure of \gls{abn} as proposed by \cite{Fukui2019Attention, Mitsuhara2019Embedding} is extensible to multiple image recognition tasks, multi-task learning problems, and various \gls{cnn} models, it is not general enough to include \gls{xai} methods other than \gls{cam}. Even though  \cite{Schiller2019Relevance, Sun2020Explanation, Zunino2021Explainable, Zunino2021Excitation} employ specific \gls{xai} methods in their experiments, any other (local) explanation technique is applicable in theory, as long as it is able to produce intermediate explanations.\\

\noindent{\bf Intermediate Feature Transformation.} 
While above attention and feature masking methods directly utilized the feature-wise information provided by \gls{xai} to determine the importance of intermediate features and to scale them accordingly, the approaches listed here exploit explanations more indirectly and rely on more complex feature transformations such as translation and projection to, e.g., correct a model's reasoning.  With this goal in mind, the \gls{clarc} framework \cite{Anders2022Finding} is formulated in a general manner and specifically aims at the identification and removal of biases, artifacts, and \gls{ch} behavior. For this purpose, the \gls{clarc} framework consists of three steps: Identifying artifacts, estimating an artifact model for each artifact, and updating the predictor model to reduce the artifact's impact. More precisely, \cite{Anders2022Finding} proposes an extension of the \gls{spray} \cite{Lapuschkin2019Unmasking} algorithm for the artifact identification, which leverages large sets of local explanations to identify a model's behavioral patterns. This condensed information can then be exploited by a human observer to pinpoint undesired biases and artifacts. Once an artifact is found and subsequently estimated, e.g., using \glspl{cav} \cite{Kim2018Interpretability}, two distinct variants of \gls{clarc} exist for removal: \gls{aclarc}, where the artifact is added to all samples regardless of class membership (with a certain probability) and the model is finetuned in order to be desensitized to the now class-unspecific artifact feature, and \gls{pclarc}, where the artifact is suppressed during inference via projection, without additionally finetuning the model. 
Both approaches can be applied in feature space as well as to the input space, since inputs can be viewed as an extension of intermediate features. In this context, the authors demonstrate that the layer where a \gls{ch} artifact can be mitigated most effectively strongly depends on the artifact's complexity. \gls{clarc} thus manages to not only identify \gls{ch} strategies, but successfully ``un-Hanses'' even large and complex datasets implicitly, such as ILSVCR2012 \cite{Russakovsky2015Imagenet},  ISIC~2019 \cite{Tschandl2018Ham, Codella2018Skin, Combalia2019Derm}, or the Adience benchmark dataset \cite{Hassner2014Age}, without sacrificing samples which might otherwise contain valuable information next to artifactual features.
Since human interaction is only required in the identification step, to interpret the model's behavioral patterns aggregated via \gls{spray}, this method is comparatively fast and relatively independent of human errors. \gls{clarc} is further extended in \cite{Pahde2022Patclarc}, by estimating artifact directions as signals, with more robustness against noise.

As demonstrated above, \gls{xai}-dependent techniques that augment intermediate features can change a model's internal feature representations in order to improve performance or achieve better reasoning. While most of these methods can boast no or only minimal requirement of human involvement --- which strongly reduces the required time and effort for their application --- they in turn cannot be model-agnostic, since they require access to a model's internal feature representations (i.e., layer-wise explanations).

\subsection{Augmenting the Loss}
\label{sec:augmentation:loss}
The loss function measures how wrong a model's predictions are. Thus, it strongly influences the decision-making that a model will learn during training,  as it controls the nature and shape of the error signal to be minimized via parameter updates. As such, augmenting the loss --- as shown in Equation~\eqref{eq:augmentation:loss:base} --- with additional regularization terms or applying a \gls{xai}-dependent scaling can naturally correct a model's reasoning, robustness, and performance, or lead to faster convergence.\\

\noindent{\bf Loss Regularization.} 
By adding a regularization term to the loss function, a model's learning behavior can easily be nudged towards a variety of effects. For instance, by comparing explanations to some ground truth that encodes human expectations, reasoning can be enforced to align with expert knowledge \cite{Ross2017Right, Liu2019Incorporating, Rieger2020Interpretations, Du2019Learning, Erion2020Improving}. Alternatively, simply imposing some human-independent constraint on the explanations may provide various positive effects, such as improved reasoning, robustness, or performance \cite{Ross2018Improving, Du2019Learning, Chen2019Robust, Erion2020Improving}.

Here, the authors of \cite{Ross2017Right} introduce \glsfirst{rrr} which aims at optimizing a model's reasoning. They assume a dataset $X$ which offers --- in addition to ground truth class labels --- a binary annotation mask $a_i^l$ for each sample $x_i \in X$ that denotes for each input dimension $\delta \in \{1, ..., D\}$, whether it should be irrelevant ($a_i^l[\delta] = 1$) to the model's decision. Note that in the original approach, $l = 0$,  so that only annotation masks $a_i^0$ in the input space are considered. The loss function can then simply be augmented by adding an additional regularization term that aims to align the explanation of each prediction with the corresponding annotation mask:
\begin{equation}
\mathcal{L}_\text{rrr}(f_{\theta^t}(x_i), y_i) = \mathcal{L}_{\text{pred}}(f_{\theta^t}(x_i), y_i) + \lambda \mathcal{L}_{\text{reason}}(r_i^{l, t}, a_i^l),
\label{eq:rrr:total_loss}
\end{equation}
where $\lambda$ is a regularization parameter, $\mathcal{L}_{\text{pred}}$ denotes the standard prediction loss term between the true and predicted class probabilities including any non-\gls{xai}-based regularization terms. In addition to learning to predict accurately through $\mathcal{L}_{\text{pred}}$, the reasoning loss term $\mathcal{L}_{\text{reason}}$ enforces correct reasoning. In their approach, the authors employ explanations $r_i^{l, t}$ at layer $l = 0$, \ie, input gradient \wrt\ log outputs $r_i^{0, t} = \partial \sum_{c=1}^{C} log(f_{\theta^t}(x_i)) / \partial x_i$ as their explanation method of choice, where large values indicate input elements that strongly affect the prediction when changed. Here, $C$ denotes the number of classes. Therefore, in order to align the model's decision-making to the ground truth input irrelevancies $a_i^l$ of each sample, input gradients should be close to zero where $a_i^l[\delta] = 1$, \ie, for irrelevant parts. This goal is captured by the reasoning loss term $\mathcal{L}_{\text{reason}}$:
\begin{equation}
\mathcal{L}_{\text{reason}}(r_i^{l, t}, a_i^l) = ||a_i^l \odot r_i^{l, t}||_2^2,
\label{eq:rrr:reason_loss}
\end{equation}
where $||\cdot||_2$ denotes the $\ell_2$-norm and $\odot$ the element-wise product. We employed the same loss term during Toy Experiment 3 in Section~\ref{sec:background:improving:toy3} with \gls{lrp} as the explanation method instead of log probabilities input gradients. Note that the $r_A$ we defined for Toy Experiment 3 defines which features should be \emph{relevant} to the model's decision and is therefore complementary to the $a_i^l$ employed here. Using this method, the authors of \cite{Ross2017Right} are able to align model reasoning with human expectations, and even have models generalize on ambiguous datasets. Although, for this purpose, annotation masks have to be provided by a human expert time-consumingly. An alternative is discussed for the case if these annotations are not provided, where multiple models are trained sequentially, each constrained to learn qualitatively different reasoning compared to the previous ones. However, this is not only quite costly due to the large amount of training processes, but a human expert is still required to select the model with a desired reasoning. 

In \cite{Ross2018Improving}, the human involvement is completely left out, by simply removing $a_i^l$ from the equation. Consequently, instead of reasoning, the robustness of models against adversarial perturbations is improved here, although the authors note that the model's decision boundaries change as a side-effect, implying a different --- although not necessarily better --- reasoning. The approaches of \cite{Ross2017Right, Ross2018Improving} are largely model-agnostic (as long as the models and explanations are differentiable) and, even though input gradients are used in the original formulation, they are easily generalizable and adaptable to other local \gls{xai} techniques. 

For instance, to reduce model bias and boost performance in scarce data settings for text classification \cite{Liu2019Incorporating} employ \gls{ig} \cite{Sundararajan2017Axiomatic} to incorporate priors into the loss function via an \gls{xai}-based regularization term. Similarly to \cite{Ross2017Right}, an attribution target, \ie, a ground truth attribution is required for each sample (and also \wrt\ each class, which is different to \cite{Ross2017Right}). However, this target annotation is not necessarily binary here, and only needs to be manually specified for samples and \wrt\ classes of interest. The loss function is simply regularized using the squared distance between each attribution and attribution target. 
Based on Contextual Decomposition \cite{Murdoch2018Beyond}, the authors of \cite{Rieger2020Interpretations} introduce \gls{cdep}, where explanations are utilized in a similar manner, regularizing by using the absolute difference between explanations and ground truth attributions. They note that obtaining ground truth explanations from human experts may be too costly, and thus propose using programmatic rules to identify important regions. The authors of \cite{Rieger2020Interpretations} demonstrate that their method can successfully correct model reasoning on various real-world datasets, such as the ISIC skin cancer classification dataset.

For \gls{vqa}, \cite{Selvaraju2019Taking} assign scalar importance scores to region proposals (\ie, image regions that the model proposes to utilize for answering a given question) by summing over the gradients of the ground truth output \wrt\ proposal features using a modified version of the feature map importance computation proposed for \gls{gcam} \cite{Selvaraju2016Gradcam}. The region proposal importance scores are then compared to human attention scores. For this purpose, the ranking of region proposal importances is enforced to be similar to the human baseline through a ranking loss term, in addition to the standard loss term promoting high task performance. In its original formulation, the approach of \cite{Selvaraju2019Taking} is restricted to their modified version of \gls{gcam} to obtain network importance scores, however, given a method to reduce arbitrary explanations to scalar importance scores (\eg, through the ratio of importance inside the proposal region and total importance, in the same manner that the authors reduce human attention maps to scalar values), any local explanation method would be applicable.

In the context of text classification, \cite{Du2019Learning} further pursue the idea of \gls{xai}-dependent loss regularization by introducing three additional regularization terms, with the respective aims of reducing the impact of irrelevant input features, making explanations uncertain if the input contains no important features, and offering sparse explanations. Here, a representation erasure based method \cite{Li2016Understanding} is applied to obtain explanations. While the first two regularization terms again require (binary) human ground truth annotations for each sample, the last regularization term is only applied to samples where this annotation is not available. Using this method, \cite{Du2019Learning} achieve models that not only have less biased reasoning, but are also able to generalize better. 

By regularizing the loss function so that the attributions themselves become more robust, \cite{Chen2019Robust} are simultaneously able to increase the robustness of the predictions a model makes, as well as align the model's reasoning more with human perception. Their experiments are based on the \gls{ig} attribution method, which is computed between an input and a reference input. By minimizing the \gls{ig} attributions between an input and references close to it in the loss function, similar attributions for similar inputs are enforced in an automated manner, thereby increasing model robustness. However, the approach of \cite{Chen2019Robust} is relatively constrained to \gls{ig} attributions, due to the specificity of the proposed loss regularization to this \gls{xai}-method.
Similarly, \cite{Ismail2021Improving} obtain higher-quality (local) explanations by reducing low gradient noise. For this purpose, they add a regularization term to the loss function that encourages similar predictions for the original inputs and inputs where the parts with low gradient values are masked. The achieved models are able to distinguish better between informative and uninformative features, resulting in clearer explanations. In contrast to many other approaches that augment the loss function, the technique proposed in \cite{Ismail2021Improving} does not require any ground truth explanations to be provided. 

The authors of \cite{Erion2020Improving} formalize the introduction of attribution priors into the loss function, thereby providing a generalized framework for \gls{xai}-based loss regularization approaches, including the ones mentioned above, and explore this framework in various concrete settings. They summarize a \gls{xai}-regularized loss function in general as
\begin{align}
\mathcal{L}_{\text{xai-reg}}(f_{\theta^t}(x_i), y_i) = \mathcal{L}_{\text{pred}}(f_{\theta^t}(x_i), y_i) + \lambda \zeta(r_i^{l, t}),
\label{eq:att_prio:general_loss}
\end{align}
where $\mathcal{L}_{\text{pred}}$ is the prediction loss between ground truth and predictions, $\lambda$ is a scalar factor, and $\zeta(r_i^{l, t})$ denotes an arbitrary scalar-valued penalty function on the attributions. This formulation is general enough to include cases where human ground truth attribution annotations are available, and cases where these annotations are not available. The authors of \cite{Erion2020Improving} demonstrate how this approach can be employed to improve model robustness, efficiency through sparsity, and performance, although their formulation is general enough allow for improving even more properties.

In order to improve models based on \gls{xai}, approaches that regularize the loss function are the most researched, as these types of methods are not only extremely versatile in the improvement goals that can be accomplished and the \gls{xai}-techniques that can be utilized, but also comparatively easy to implement and inherently model-agnostic --- apart from the specific \gls{xai}-methods potentially requiring access to internal model parameters. 
In terms of efficiency, there is a high variance between approaches: When gradient-based attributions are used during training in the regularization term, including modified backpropagation approaches, second-order gradients need to be computed to optimize the loss \cite{Rieger2020Interpretations}, which moderately increases computational costs. The inclusion of attributions into the loss function further imposes the restrictions that only explanation methods that are differentiable \wrt\ the model weights can be employed, and that the model needs to be twice differentiable (a condition which most current \gls{dnn} architectures fulfill). Some approaches require ground truth attribution annotations by human experts, often for each single sample, which are time-consuming to obtain, or even infeasible for large datasets. However, some human guidance may be required for improvement goals such as reasoning, since valid and invalid input features are generally impossible to determine based on the available training data only. Note also, that \cite{Dombrowski2019Explanations, Anders2020Fairwashing} showed that explanations can potentially be manipulated so that they do not necessarily reflect the actual classifier behavior. Thus, a model that is trained via an \gls{xai}-regularized loss in order to correct its reasoning could theoretically simply learn to emulate the correct explanations while still basing its predictions on invalid input features.\\

\noindent{\bf Loss Scaling.} While above methods introduce a regularization term into the loss function in order to achieve various effects, in the case of imbalanced data, the class-wise losses can simply be scaled based on information offered by \gls{xai}. The approach of \cite{Weber2020Towards} --- which is already discussed in-depth in Section~\ref{sec:augmentation:data} --- employs class-wise factors based on \gls{lrp} (although other local \gls{xai} methods can in theory be substituted) for this very purpose. As a result, the model's convergence speed is improved in an automated manner. 
Similarly to regularizing the loss, scaling losses does not require any access to the model's internal parameters. This approach is, however, much less general, since a variety of model properties can be controlled via an added regularization term, but simple scaling factors on the loss are only able to change training behavior in a comparatively limited fashion.

\subsection{Augmenting the Gradient}
\label{sec:augmentation:gradient}
The gradient determines the direction and speed with which a model's parameters are updated during the backward pass. By modifying either the intermediate feature gradients, or the parameter gradients directly, as described in Equations~\eqref{eq:augmentation:feature-gradient:base} and \eqref{eq:augmentation:parameter-gradient:base}, the backward flow of weight updates can be controlled, improving convergence behavior and performance.
Compared to the intermediate feature masking approaches presented in Section~\ref{sec:augmentation:feature}, where a mask is obtained from intermediate explanations in order to weigh \emph{features} during the \emph{forward pass}, a similar mask can be computed based on \gls{xai}, denoting the importance of \emph{gradients} during the \emph{backward pass}. While all gradient transformation approaches seek to alter the ratio with which model parameters are updated, two distinct types of methods can be distinguished here:

Due to their shape being the same as the explained intermediate features, explanations can directly be employed to obtain importance scores of the feature gradients $\frac{\partial \mathcal{L}(f_{\theta^t}(X), Y)}{\partial f^l_{\theta^t}(X)}$, see  Equation~\eqref{eq:augmentation:feature-gradient:base} (\emph{top}), and augment them accordingly. By doing this, parameter updates of all layers below the augmentation are affected, as is the case for, \eg, \cite{Nagisetty2020xai}. 
Alternatively, parameter gradients $\frac{\partial \mathcal{L}(f_{\theta^t}(X), Y)}{\partial \theta^{l, t}}$, see Equation~\eqref{eq:augmentation:parameter-gradient:base} (\emph{bottom}), can be augmented directly based on \gls{xai}. In order to do this, weight-wise importance scores first need to be obtained from intermediate explanations, which is not always trivial due to the shape mismatch, but nevertheless employed by, \eg, \cite{Lee2019Improvement}. In contrast to augmenting the feature gradients, only the weight update at the augmented layer is affected here.

In the context of \glspl{gan}, \cite{Nagisetty2020xai} propose \gls{xai}-\gls{gan}, which improves upon the generator optimization by augmenting the feature gradients at a singular intermediate layer. In standard \glspl{gan}, the generator $g$ is tasked with fooling the discriminator $d$, so that, given a noise sample $z_i, i \in \{1, ..., N\}$, a generated sample $g_{\theta_g^t}(z_i)$ cannot be distinguished by $d$ from real samples $X$. Usually, the generator and discriminator are trained in an alternating fashion, with the parameters $\theta_g^t$ of $g$ being updated by backpropagating the loss of the discriminator prediction $d_{\theta_d^t}(g_{\theta_g^t}(z_i))$ through the discriminator, yielding the gradient of the generated example $\nabla_{g(z_i)} = \frac{\partial \mathcal{L}(d_{\theta_{d}^{t}}(g_{\theta_g^t}(z_i)), y_i)}{\partial g_{\theta_g^t}(z_i)}$. The generator gradients $\nabla_{g}$ are then simply computed from $\nabla_{g_{\theta_g^t}(z_i)}$, in order to update the generator weights. However, \cite{Nagisetty2020xai} suggest to first augment $\nabla_{g_{\theta_g^t}(z_i)}$ with a importance mask $M^t$, which is obtained from the explanation of the discriminators prediction:
\begin{equation}
\nabla_{g_{\theta_g^t}(z_i)}' = \nabla_{g_{\theta_g^t}(z_i)} + \lambda \nabla_{g_{\theta_g^t}(z_i)} M^t, 
\label{eq:xaigan:update}
\end{equation}
where $M^t$ denotes the importance of each feature of $g_{\theta_g^t}(z_i)$ towards the disciminator's decision, and $\lambda$ is a scalar hyperparameter. 

In contrast, \cite{Lee2019Improvement} propose various variants of parameter gradient augmentation. Based on \gls{lrp}, the authors compute an attribution $r_i^{l, t}$ for each layer $l$ in a model. For two layers $l$ and $l+1$, with $M$ and $N$ neurons, respectively, and corresponding attribution maps $r_i^{l, t} \in \mathbb{R}^{M}$ and $r_i^{l+1, t} \in \mathbb{R}^{N}$, weight-wise attribution scores are then obtained via a simple dot product, solving the shape mismatch problem discussed above as follows (note that this equation would require $M = N$):
\begin{equation}
r^{l, t}_{\text{w}, i} = r_i^{l, t} \cdot r_i^{l+1, t} \qquad .
\label{eq:lrpimprov:Rw}
\end{equation}
Then, $r^{l, t}_{\text{w}, i}$ is simply normalized into the interval $[0, 1]$ by division through its sum, yielding $(r_{\text{w}, i}')^{l, t}$, and can be employed to augment the gradient, \eg, by using weight-wise learning rates for a simple multiplication:
\begin{equation}
(\theta')^{l, t} = \theta^{l, t} - \eta \frac{\partial \mathcal{L}(f_{\theta^t}(x_i), y_i)}{\partial \theta^{l, t}} r^{l, t}_{\text{w}, i}~,
\label{eq:lrpimprov:weights}
\end{equation}
where $\eta$ denotes the scalar learning rate. The authors of \cite{Nagisetty2020xai}  demonstrate the variability \wrt\ the employed \gls{xai} technique by using their method of feature gradient augmentation with \gls{lime},  DeepSHAP \cite{Lundberg2017Unified}, or DeepLIFT \cite{Shrikumar2017Learning}. They obtain generator models that do not only produce higher quality samples, but do so much more efficiently, and can therefore be trained using only a fraction of the data compared to standard \glspl{gan}. The authors of \cite{Lee2019Improvement} suggest multiple variations of their technique based on the weight-wise importance mask. They test their method on various image classification datasets, and report similar or better performance and convergence --- although not necessarily faster convergence, as well as higher confidence of the resulting models, compared to some state-of-the-art optimization schemes such as \gls{sgd} \cite{Bottou2012Stochastic} and Adadelta \cite{Zeiler2012Adadelta}. However, their solution to obtaining weight-wise importance scores is based on a dot product between the explanations of consecutive layers --- and as such is not applicable if these consecutive layers are not fully-connected or do not contain the same number of neurons. Using the outer product instead would allow for applying this method to dense layers of arbitrary neuron numbers, although other layer types are still not considered. Both methods can in theory be applied using any \gls{xai} technique that is able to provide layer-wise explanations.

By not considering all gradients equally during backpropagation and instead using explanations to select the most important ones, \gls{xai}-based gradient augmentation methods are able to affect the direction of each learning step, leading to distinct effects, such as better performance and convergence, or even increased data efficiency. Since most of these methods do not require any human involvement, they can easily be employed in order to achieve better models without requiring significantly more training time. However, these methods are generally not model-agnostic, simply because access to a model's internal parameters is required by definition.

\subsection{Augmenting the Model}
\label{sec:augmentation:model}
Most of the above augmentation categories are employed during training in order to improve the model. However, even after obtaining a good model, some undesirable properties may persist, such as a large number of parameters and thus the large required disc space and high computational effort. These properties depend on the model definition itself and can thus best be improved upon by augmenting the model, as described by Equation~\eqref{eq:augmentation:model:base}.

Algorithms for pruning or quantization rely on an estimation of each parameter's importance to the model's decision-making and performance. As such, the information offered by \gls{xai} naturally presents such a criterion, and can therefore be leveraged in order to increase model efficiency in terms of required computational cost of inference or storage space. 
For this purpose, \gls{xai}-methods that can provide layer-wise explanations are required. Based on these, the connectivity and parameters of a model are altered. To prune a trained model, existing approaches \cite{Yeom2019Pruning, Sabih2020Utilizing} compute intermediate attributions, \eg, for a small number of reference samples, and average these to obtain a pruning criterion in the form of importance scores. Consequently, the neurons or filters with the lowest importance are pruned first, yielding a more storage-efficient model. Similarly, the same importance scores can be leveraged in order to quantize the model's weights, and thus reduce their memory requirements. 
With this methodology, \cite{Yeom2019Pruning} propose a pruning criterion based on \gls{lrp}, which offers intermediate explanations at any layer of a given model. They compute attributions \wrt\ the respective true class of each sample, and obtain their importance-score based pruning criterion by averaging the (absolute) relevances over the reference samples. Employing this method, the authors of \cite{Yeom2019Pruning} are able to successfully remove unimportant units while preserving performance, both when fine-tuning and not fine-tuning the model after pruning, and, in the latter case, vastly outperform other state-of-the-art approaches \cite{Sun2017meProp, Molchanov2016Pruning, Li2017Pruning}.

Similarly, \cite{Sabih2020Utilizing} utilize DeepLIFT to obtain importance scores for neural network pruning. After first normalizing these scores using the $l_1$-norm for each layer to be considered, the neurons with the lowest importance scores can then be pruned accordingly. Additionally, \cite{Sabih2020Utilizing} propose to employ the DeepLIFT scores in order to quantize a model. Here, either weight sharing is applied by k-means clustering weights \cite{Hartigan1979Kmeans} using a DeepLIFT-weighted mean squared error, where all weights within a cluster then share the same value for quantization, or mixed-precision integer quantization is employed, where the bit-precision of weights is reduced depending on their DeepLIFT scores, leading to a significant boost to storage efficiency.

The authors of \cite{Becking2021Ecq} employ explainability in order to improve upon \gls{ecq} (a generalization of EC2T \cite{Marban2020Learning}). As a clustering-based quantization algorithm, \gls{ecq} not only considers the distance of each weight to each centroid, but additionally promotes sparse assignments due to an entropy term based on the fractions of weights close to each centroid. However, this may also lead to significant model degradation if important weights are assigned to zero, as weight magnitude does not always indicate importance. By additionally considering weight-wise importance scores obtained through \gls{lrp} for the zero-centroid, \cite{Becking2021Ecq} are able to increase model storage efficiency by generating low bit width and sparse networks, and simultaneously preserving or even improving performance.

In contrast to the above approaches, the authors of \cite{Ha2021Adaptive} utilize \gls{xai} for transferring knowledge, building a completetly different model with beneficial properties and similar behavior instead of altering a singular model. More specifically, their technique, \gls{awd}, transfers knowledge from a pre-trained \gls{dnn} into a learnable wavelet transform. This is achieved by integrating an interpretation loss term into the optimization problem for obtaining the wavelet functions, based on \gls{trim} \cite{Singh2020Transformation} and saliency \cite{Simonyan2014Deep}, that ensures sparse attributions in the space of wavelet coefficients, \ie, forcing the wavelet transformation to encode model predictions as concisely as possible. 

Since computing \gls{lrp} and DeepLIFT only requires a modified backward pass while saliency simply requires computing gradients, and since no human interaction is required, the approaches of \cite{Yeom2019Pruning}, \cite{Sabih2020Utilizing}, and \cite{Becking2021Ecq} are computationally extremely efficient. However, the inherent optimization problem for \gls{awd} \cite{Ha2021Adaptive} requires evaluation over multiple data points and coefficients in the wavelet representation, making it comparatively expensive. As the authors of \cite{Yeom2019Pruning} note, \gls{lrp}-based pruning can improve upon both computational inference cost and model storage requirements, depending on whether the pruning focuses on convolutional or dense layers, respectively. Using their proposed methods for \gls{xai}-based pruning and quantization, the authors of \cite{Sabih2020Utilizing} are able to optimize memory usage and latency in an automated fashion, and apply their \gls{xai}-based criterion to a wide range of pruning and quantization variations. By including \gls{xai} into existing quantization techniques, the authors of \cite{Becking2021Ecq} are able to significantly reduce the trade-off between efficiency and performance. The authors of \cite{Ha2021Adaptive} show that their resulting wavelet transformations lead to far smaller, simplified, more computationally efficient, and more inherently interpretable models, while simultaneously preserving performance. While above approaches employ one specific \gls{xai} method each, any explanation technique that is able to provide intermediate attributions can be substituted in theory, although this may affect computational efficiency and quality of the resulting pruning or quantization criterion.

By changing a model's parameters based on the information offered by \gls{xai}, model efficiency in terms of storage cost and inference speed is easily increased. Similarly to \gls{xai}-dependent intermediate feature augmentation methods, above model augmentation techniques require layer-wise explanations, and are by definition not model-agnostic, as access to the model's parameters is required. However, they are applicable in an automated fashion, and often do not even need any finetuning, making their usage effortless and time-efficient.

\subsection{Approaches not Considered}
\label{sec:augmentation:notconsidered}
In this review, we only focus on model improvement methods where \gls{xai} is employed to achieve that goal. Of course the research community investigating ways to design and train more performant, reliable and efficient models is much broader. For instance, approaches such as \cite{Zaidan2007Using, Zaidan2008Modeling, Hendricks2018Women, Zhang2016Rationale, McDonnell2016Why} utilize human knowledge to providing better annotations, and thus achieve better performing models that make decisions for better reasons. While not necessarily providing additional information beyond labels, techniques from the field of human-machine interaction \cite{Tong2001Support, Judah2012Active, Shivaswamy2015Coactive, Gal2017Deep} also leverage external knowledge, but expect a (potentially human) expert to actively interact with the model. This may slow down the learning process but allows for specific, isolated, and more targeted labels or corrections to be given in order to learn with a lower sample complexity \cite{Balcan2010The}. These approaches which rely on human knowledge only (i.e., do not employ \gls{xai}) are not considered here. 
Also the whole research field aiming at training models that inherently explain themselves \cite{Rudin2019Stahp, Chen2019This, Wu2019Towards, Chen2020Concept, Barnett2021Interpretable} is out of scope of this review. This can also be considered an improvement, since explainable models do not necessarily trade-off in terms of performance and may even have more desirable robustness and reasoning properties \cite{Rudin2019Stahp}. Nevertheless, these approaches tend to make significant restrictions to the model architecture, and they do not employ \gls{xai} to improve models, but instead improve the model's explainability itself.

\subsection{Discussion}
\label{sec:augmentation:discussion}
The various previously described approaches for \gls{xai}-based model improvement not only differ in terms of the model property that is improved, but also \wrt\ what exactly is augmented, \ie, the data distribution, the intermediate features during the forward pass, the loss function, the gradients during the backward pass, or the model after training. An overview over this two-dimensional categorization is shown in Table~\ref{tab:improvement_augmentation_types}. Note that this table is quite sparse, suggesting that each type of augmentation is only suited for improving specific model properties. \gls{xai}-based loss augmentation seems to be extremely versatile, being able to improve model performance, robustness, efficiency, and reasoning. This is not surprising, since many of the above properties can be simply expressed as an additional regularization term. On the other hand, \gls{xai}-based model augmentation approaches, which are generally applied after training, and include short finetuning at most, only seem to improve model efficiency --- albeit with remarkable success \cite{Yeom2019Pruning, Becking2021Ecq}. Other properties, which are generally decided during training, therefore do not seem influenceable through model augmentation.

\renewcommand{\arraystretch}{2}
\begin{table}[!ht]
    \centering
    \caption{Overview and Categorization of Approaches that aim to improve \gls{ml}-models using \gls{xai}.}
    \renewcommand{\arraystretch}{1}
    \setlength\tabcolsep{1pt}
    \hspace*{-1cm}
    \begin{tabular}{c}
    \resizebox{\textwidth}{!}{
    \begin{tabular}{l|C|C|C|C|C|C|C|C|C|C|C|C|C|C|C|C|C|C|C|C|C|C|C|C|C|C|C|C}
           \multicolumn{1}{c}{}
         & \multicolumn{5}{c}{\cellcolor{pastelviolet}}%
         & \multicolumn{1}{c}{\cellcolor{pastellightred}}
         & \multicolumn{6}{c}{\cellcolor{pastelblue}}%
         & \multicolumn{1}{c}{\cellcolor{pastellightblue}}%
         & \multicolumn{9}{c}{\cellcolor{pastelred}}%
         & \multicolumn{2}{c}{\cellcolor{pastelorange}}%
         & \multicolumn{4}{c}{\cellcolor{pastelgreen}}%
         \\
         & \cite{Teso2019Explanatory} %
         & \cite{Schramowski2020Making} 
         & \cite{Gautam2021This}
         & \cite{Bargal2019Guided}%
         & \multicolumn{2}{c|}{\cite{Weber2020Towards}}
         & \cite{Fukui2019Attention} %
         & \cite{Mitsuhara2019Embedding} 
         & \cite{Schiller2019Relevance} 
         & \cite{Sun2020Explanation} 
         & \cite{Zunino2021Explainable} 
         & \cite{Zunino2021Excitation} 
         & \cite{Anders2022Finding}%
         & \cite{Ross2017Right} %
         & \cite{Ross2018Improving} 
         & \cite{Liu2019Incorporating} 
         & \cite{Rieger2020Interpretations} 
         & \cite{Selvaraju2019Taking}
         & \cite{Du2019Learning} 
         & \cite{Chen2019Robust}
         & \cite{Ismail2021Improving}
         & \cite{Erion2020Improving} 
         & \cite{Nagisetty2020xai} %
         & \cite{Lee2019Improvement} 
         & \cite{Yeom2019Pruning} %
         & \cite{Sabih2020Utilizing} 
         & \cite{Becking2021Ecq} 
         & \cite{Ha2021Adaptive}
         \\
        \noalign{\hrule height 0.5pt}
        Performance 
        & %
        & &
        & \cellcolor{pastelviolet} X
        & &
        & \cellcolor{pastelblue} X %
        & \cellcolor{pastelblue} X 
        & \cellcolor{pastelblue} X 
        & \cellcolor{pastelblue} X 
        & \cellcolor{pastelblue} X 
        & \cellcolor{pastelblue} X 
        & %
        & %
        &
        & \cellcolor{pastelred} X 
        & \cellcolor{pastelred} X 
        & \cellcolor{pastelred} X 
        & \cellcolor{pastelred} X 
        & 
        & \cellcolor{pastelred} X
        & \cellcolor{pastelred} X 
        & \cellcolor{pastelorange} X %
        & \cellcolor{pastelorange} X 
        & %
        & & &
        \\ \noalign{\hrule height 0.5pt}
        Convergence 
        & %
        & & & &
        & \cellcolor{pastellightred} X 
        & %
        & & & & & 
        & %
        & %
        & & & & & & & &
        & %
        & \cellcolor{pastelorange} X
        & %
        & &  &
        \\ \noalign{\hrule height 0.5pt}
        Robustness 
        & %
        & & & & &
        & %
        & & & & & 
        & %
        & %
        & \cellcolor{pastelred} X 
        & & & & 
        & \cellcolor{pastelred} X 
        & 
        & \cellcolor{pastelred} X 
        & %
        & 
        & %
        &  &  &
        \\ \noalign{\hrule height 0.5pt}
        Efficiency 
        & %
        & & & & &
        & %
        & 
        & \cellcolor{pastelblue} X 
        & & &
        & %
        & %
        & & & & & & & 
        & \cellcolor{pastelred} X 
        & \cellcolor{pastelorange} X %
        & 
        & \cellcolor{pastelgreen} X %
        & \cellcolor{pastelgreen} X 
        & \cellcolor{pastelgreen} X 
        & \cellcolor{pastelgreen} X 
        \\ \noalign{\hrule height 0.5pt}
        Reasoning 
        & \cellcolor{pastelviolet} X %
        & \cellcolor{pastelviolet} X 
        & \cellcolor{pastelviolet} X 
        & \cellcolor{pastelviolet} X 
        & &
        & %
        & \cellcolor{pastelblue} X 
        & & 
        & \cellcolor{pastelblue} X 
        &
        & \cellcolor{pastellightblue} X %
        & \cellcolor{pastelred} X %
        & \cellcolor{pastelred} X 
        & \cellcolor{pastelred} X 
        & \cellcolor{pastelred} X 
        & \cellcolor{pastelred} X 
        & \cellcolor{pastelred} X
        & \cellcolor{pastelred} X 
        & & &
        & %
        & 
        & %
        & & 
        \\ \noalign{\hrule height 0.5pt}
        Equality 
        & %
        & & &
        & \cellcolor{pastelviolet} X 
        &
        & %
        & & & & &
        & %
        & %
        & & & & & & & & 
        & %
        &
        & %
        & & &
        \\ \noalign{\hrule height 0.5pt}
      \end{tabular}
      }
      \setlength\tabcolsep{6pt}
      \renewcommand{\arraystretch}{2}
      \\
      \\
        \fbox{\begin{tabular}{ll}
        \textcolor{pastelviolet}{$\blacksquare$} & Data Re-distribution \\
        \textcolor{pastelblue}{$\blacksquare$} & Attention and Intermediate Feature Masking \\
        \textcolor{pastellightblue}{$\blacksquare$} & Intermediate Feature Transformation \\
        \textcolor{pastelred}{$\blacksquare$} & Loss Regularization \\
        \textcolor{pastellightred}{$\blacksquare$} & Loss Scaling \\
        \textcolor{pastelorange}{$\blacksquare$} & Gradient Transformation \\
        \textcolor{pastelgreen}{$\blacksquare$} & Model Pruning and Quantization \\
        \end{tabular}}
    \end{tabular}
    \hspace*{-1cm}
    \label{tab:improvement_augmentation_types}
\end{table}
\renewcommand{\arraystretch}{1}

As suggested by Table~\ref{tab:improvement_augmentation_types}, \gls{xai} can be employed to improve upon a multitude of model properties. Explanations are able to measure the importance of parts of the input or intermediate features towards a model's decision --- and can therefore be viewed as an additional and high-dimensional measurement for the discussed properties, depending on the application. They thus allow for a better (compared to just relying on the prediction error) control of the model behaviour.
However, while the chances in leveraging \gls{xai} to improve \gls{ml}-models seem extremely promising, there also exist a number of pitfalls and limitations: The quality of the desired improvement to the model directly depends on the quality of the explanation itself.  If the explainer does not provide the expected information about the model in a sufficient manner, improving model properties may be impossible. Furthermore, if augmentations are applied during training, there is a danger of the model overfitting on uninformative explanations. More concisely, some augmentations (\eg, intermediate feature or gradient augmentations) rely on explanations as a feature importance measure during training. However, if the model is untrained and makes random decisions, the explanations may not be meaningful in terms of important intermediate features, and thus throw off the model's learning trajectory \cite{Sun2020Explanation}. Some model properties, \eg, correct reasoning, cannot be improved via \gls{xai} alone, because valid input features and unwanted biases can often not be distinguished from the limited domain of the dataset alone and may be subjective. In this case, the ground truth needs to be provided by humans, to be compared to the \gls{xai}-``measurements'', making such methods often extremely time-consuming and costly, up to the point of infeasibility for large datasets. Moreover, \cite{Anders2020Fairwashing, Dombrowski2019Explanations} showed that explanations can be manipulated while keeping predictions intact. This is especially dangerous for loss regularization approaches, where a minimization is directly performed using the explanation, potentially enabling the model to learn to emulate explanations while still predicting for the wrong reasons.

Nevertheless, if above limitations are kept in mind, \gls{xai} can be leveraged to improve current \gls{ml}-models significantly.

\section{Demonstrative Examples}
\label{sec:demoexamples}
The approaches discussed in Section~\ref{sec:augmentation} are quite heterogeneous in terms of improved property, augmented location, human involvement, and degree of improvement. In the following section, we aim to investigate how and under which conditions \gls{xai} can be employed to improve specific properties of models trained on highly complex datasets, as well as potential caveats and drawbacks. For this purpose, we focus --- as an example --- on the methods introduced in \cite{Weber2020Towards} (data augmentation) and \cite{Sun2020Explanation} (intermediate feature augmentation). Finally, we derive recommendations and lessons learned when employing \gls{xai} to improve \gls{ml}-models.

\subsection{Example 1 (Model Performance)}
\label{sec:demoexamples:performance}

In the following, we consider and evaluate the usage of methods from \gls{xai} during the training step of few-shot classifiers with the aim to improve the model accuracy at test time. For this purpose, we employ the method of \cite{Sun2020Explanation} (described in Section \ref{sec:augmentation:feature}) and extend it by using several local \gls{xai}-methods, in addition to \gls{lrp}. The few-shot learning setup was chosen because, firstly, it emulates the human ability to transfer learned knowledge to unseen tasks, and, secondly, it is usually employed with a small number of samples for each seen class. It is of interest for applications in problem settings with a large number of classes with only few samples available, such as tagging of images or text streams, where it could be applied to rare single tags or sets of tags, and, generally, problems with a large number of distinct configurations.

The goal of this experiment is to evaluate the general suitability of \gls{xai} to improve the accuracy of models when faced with few sample settings and out-of-domain data. 
We took two models from \cite{Sun2020Explanation}, the RelationNet \cite{RN:sung2018learning} and a graph-based few-shot classification model \cite{FEWGNN:garcia2018fewshot}, and extended them with several approaches from explainable AI, namely Guided Backpropagation \cite{Springenberg2014Striving}, Gradient$\times$Input, \gls{gcam} and Guided \gls{gcam} for the RelationNet, and Guided Gradient$\times$Input for the model from \cite{FEWGNN:garcia2018fewshot}, \gls{lrp} and the natively trained base models. We applied each explanation method at the same place, as with \cite{Sun2020Explanation}, that is from the classifier output to the input to the relation module for the RelationNet and from the classifier output to the input to the graph classification module for the model from \cite{FEWGNN:garcia2018fewshot}. When speaking of applying Gradient$\times$Input and \gls{gcam}, we used as input the input features for the aforementioned modules and the gradient until these input features. We computed explanations at training time for each class which has a prediction probability above the inverse of the number of classes which is the threshold for guessing in a few-shot problem. This was done to be consistent with the backpropagation of \gls{lrp} in \cite{Sun2020Explanation}. At test time we used the saved prediction models and computed predictions without explanations. 
We trained each model on the mini-ImageNet dataset \cite{miniImageNet:vinyals2016matching} and evaluated its few-shot accuracy at test time on mini-ImageNet and four out-of-domain datasets, Stanford Cars \cite{Krause20133D}, CUB-2011 birds \cite{Wah2011CUB}, Places \cite{Places:zhou2017places} and Plantae \cite{Plantae:van2018inaturalist}. We used the same training parameters as in \cite{Sun2020Explanation}, however, repeated training with five different random seeds and report the average over those, as well as the standard deviation over the five runs.

\begin{table}[]
    \centering
    \caption{Performance of RelationNet (\emph{Top}) and Graph Classification-based model (\emph{Bottom}) trained for few-shot classification with and without XAI-based support. In each cell, the mean and standard deviation of the accuracy over five seeds is reported. For Gradient$\times$Input and RelationNet two of the seeds diverged, likely due to too high learning rate. The average and standard deviation of the three undiverged seeds are reported in brackets. The \textbf{bold} numbers indicate the highest mean measurement of the row, the \underline{underlined} numbers the second highest. The last row counts the number of settings in which the respective \gls{xai}-methods improved upon the baseline.}
    \begin{tabular}{l|cccccc}
    Dataset & Baseline & LRP & GBP & Grad$\times$Inp & GradCam & GuiGrCAM\\ \hline
    Cars     &\underline{39.10}$\pm$0.69 & \textbf{39.56}$\pm$0.86 &37.31$\pm$1.15 & 31.35$\pm$10.38& 37.96$\pm$1.17& 37.96$\pm$1.17 \\
    &&&&(38.92$\pm$0.79)&&\\
    CUB     &57.69$\pm$1.15 &\textbf{58.27}$\pm$0.44 &56.49$\pm$1.37 & 42.88$\pm$20.88& 57.08$\pm$1.66& \underline{58.00}$\pm$0.71\\ 
    &&&&(58.13$\pm$0.40)&&\\
    MiniImg   &72.25$\pm$0.23 & \textbf{72.65}$\pm$0.55 &71.69$\pm$0.72 &51.66$\pm$28.90 &72.38$\pm$0.76 &\underline{72.63}$\pm$0.51 \\ 
    &&&&(72.76$\pm$0.70)&&\\
    Places   &64.75$\pm$0.36 &64.74$\pm$0.12 &63.95$\pm$0.74 &47.03$\pm$24.68 &\underline{64.80}$\pm$1.02 & \textbf{64.98}$\pm$0.70\\ 
    &&&&(65.05$\pm$0.24)&&\\
    Plantae  &\underline{47.15}$\pm$1.01 &\textbf{48.07}$\pm$0.84 & 45.26$\pm$0.75 &35.77$\pm$14.39 & 46.22$\pm$1.14 &46.77$\pm$0.64 \\
    &&&&(46.28$\pm$0.27)&&\\
    Better than Base & -- & 4 & 0 & 0 & 3 & 3 \\ 
    &&&&(3)&&\\ \hline
    \end{tabular}
    \label{tab:fewshot_rn}
\end{table}  

\begin{table}[]
    \centering
    
        \begin{tabular}{l|ccccc}
    Dataset & Baseline & LRP & GBP & Grad$\times$Inp  & GuiG$\times$Inp\\ \hline
    Cars    &44.95$\pm$1.16 &\underline{45.50}±0.70 &44.28±0.61 &\textbf{45.56}±0.82  &45.36±0.64\\ 
    CUB     &63.86±1.74 &\underline{64.96}±1.46 &64.65±1.19 &\textbf{65.19}±0.35 &64.88±1.00 \\ 
    MiniImg  &80.33±0.83 &80.40±0.29 &80.02±0.92 &\textbf{81.30}±0.92 &\underline{80.71}±0.34\\ 
    Places   &70.38±1.17& \underline{72.09}±0.51 &71.06±0.51 &\textbf{72.63}±1.07 &71.59±0.60\\ 
    Plantae &55.60±1.02 &\textbf{57.30}±0.96 & 55.86±0.94 &\underline{56.04}±1.37 &55.80±0.72\\
        Better than Base & -- & 5 & 3 & 5 & 5  \\\hline
    \end{tabular}
\end{table}

As shown in Table \ref{tab:fewshot_rn}, model performance in the few-shot context can be reliably improved through application of \gls{xai}. Although the increase in accuracy is often only marginal (\ie, $<1$\%), this observation consistently holds with only minor variations across datasets, models, and seeds (\eg, cf. results for \gls{lrp}). However, there are considerable differences between explanation techniques. For instance, \gls{lrp} and Gradient$\times$Input (if the diverged seeds are disregarded) lead to significant improvements in the majority of settings, \ie, in 9/10 and 8/10 cases, respectively. On the other hand, \eg, Guided Backpropagation only leads to a better performance in 3/10 of the considered settings. This may be due to the fact that Guided Backpropagation is the only of the applied \gls{xai}-methods that does not take intermediate features into account, and instead only expresses a modified sensitivity (which --- except for the influence of ReLUs being turned on or off --- exclusively relies on the model parameters). In any case, the reasons behind the observed differences in suitability of \gls{xai} methods for model improvement are subject to further research.

\subsection{Example 2 (Model Equality)}
\label{sec:demoexamples:equality}

\begin{figure}[!ht]
    \centering
    \includegraphics[width=0.8\linewidth]{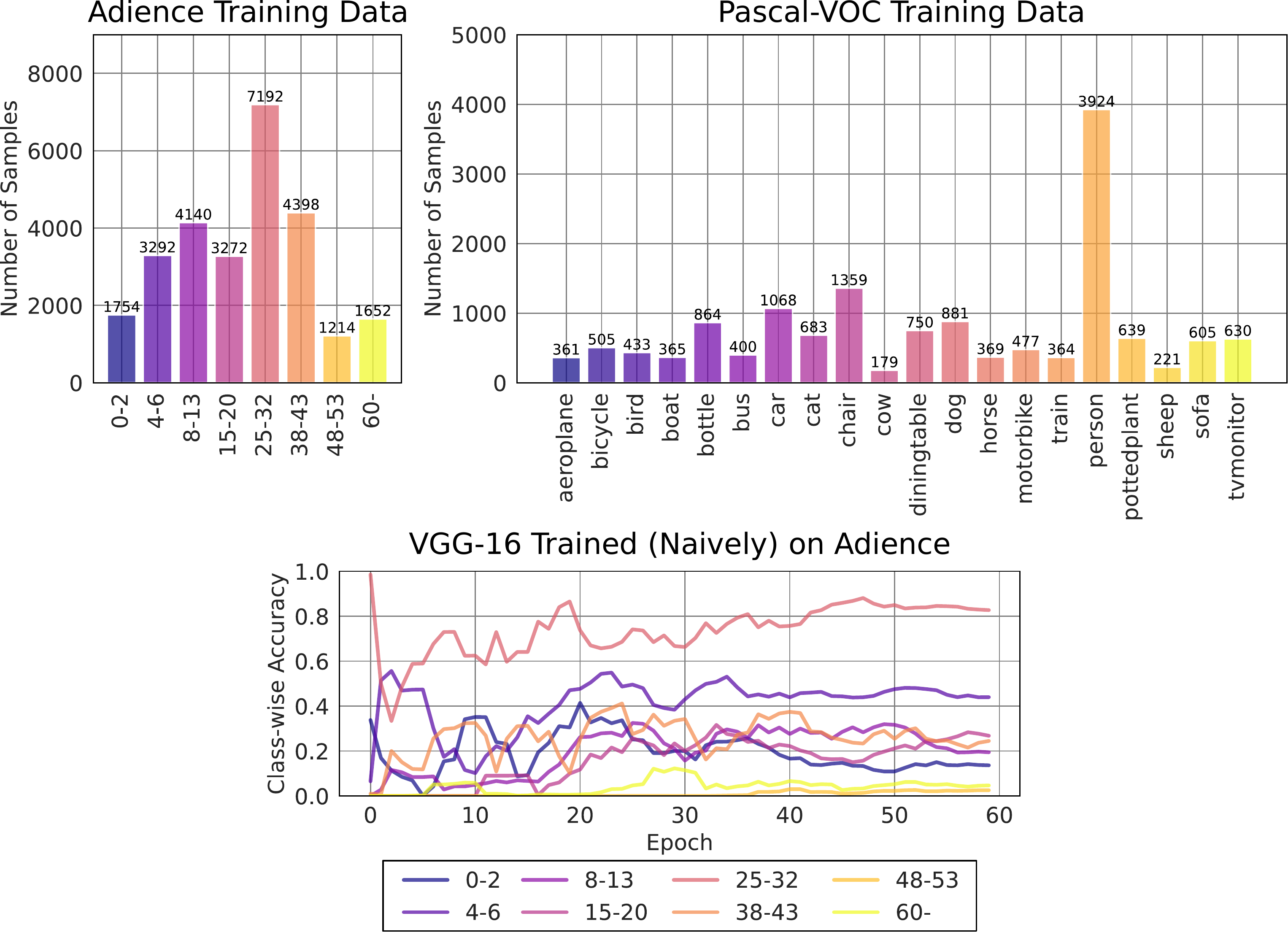}
    \caption{Visualization of class-wise (training set) distributions of the Adience~\cite{Hassner2014Age} benchmark dataset of unfiltered faces (\emph{top left}) and the Pascal VOC 2012 challenge dataset~\cite{Everingham2012Pascalvoc} (\emph{top right}). If the available data is this imbalanced, models trained naively on it easily overfit on majority classes while ignoring minority classes (\emph{bottom}).}
    \label{fig:datadistributions}
\end{figure}

Datasets that reflect realistic circumstances --- such as the Adience~\cite{Hassner2014Age} benchmark dataset of unfiltered faces or the Pascal VOC 2012 challenge dataset~\cite{Everingham2012Pascalvoc} --- often do not contain the same amount of examples for each class. 
If this imbalance is too significant, it can lead to models trained on those datasets overfitting on majority classes while largely ignoring minority classes, since modern \gls{ml}-systems tend to minimize error criteria estimated over \emph{all} available data, without class-wise distinctions. For reference, the extremely imbalanced distribution of samples per class for Adience and Pascal-VOC is visualized in Figure~\ref{fig:datadistributions} (\emph{top}), which easily results in a skewed class-wise performance over training of a naively trained model (\emph{bottom}).  

However, as shown by \cite{Weber2020Towards}, the class-wise performance of a model can be estimated through explanations (here specifically \gls{lrp}), even before any changes in class-wise performance metrics are detectable. In the following, we employ their method of \gls{xai}-guided imbalance mitigation to re-balance the distribution of input data based on the entropy of and MSE-distance between explanations. Details on the data and models used, on the employed explanation and augmentation methods, and about the training scheme and evaluation metrics can be found in \ref{sup:sec:demodetails:equality}. %

\begin{figure}[!ht]
    \centering
    \includegraphics[width=\linewidth]{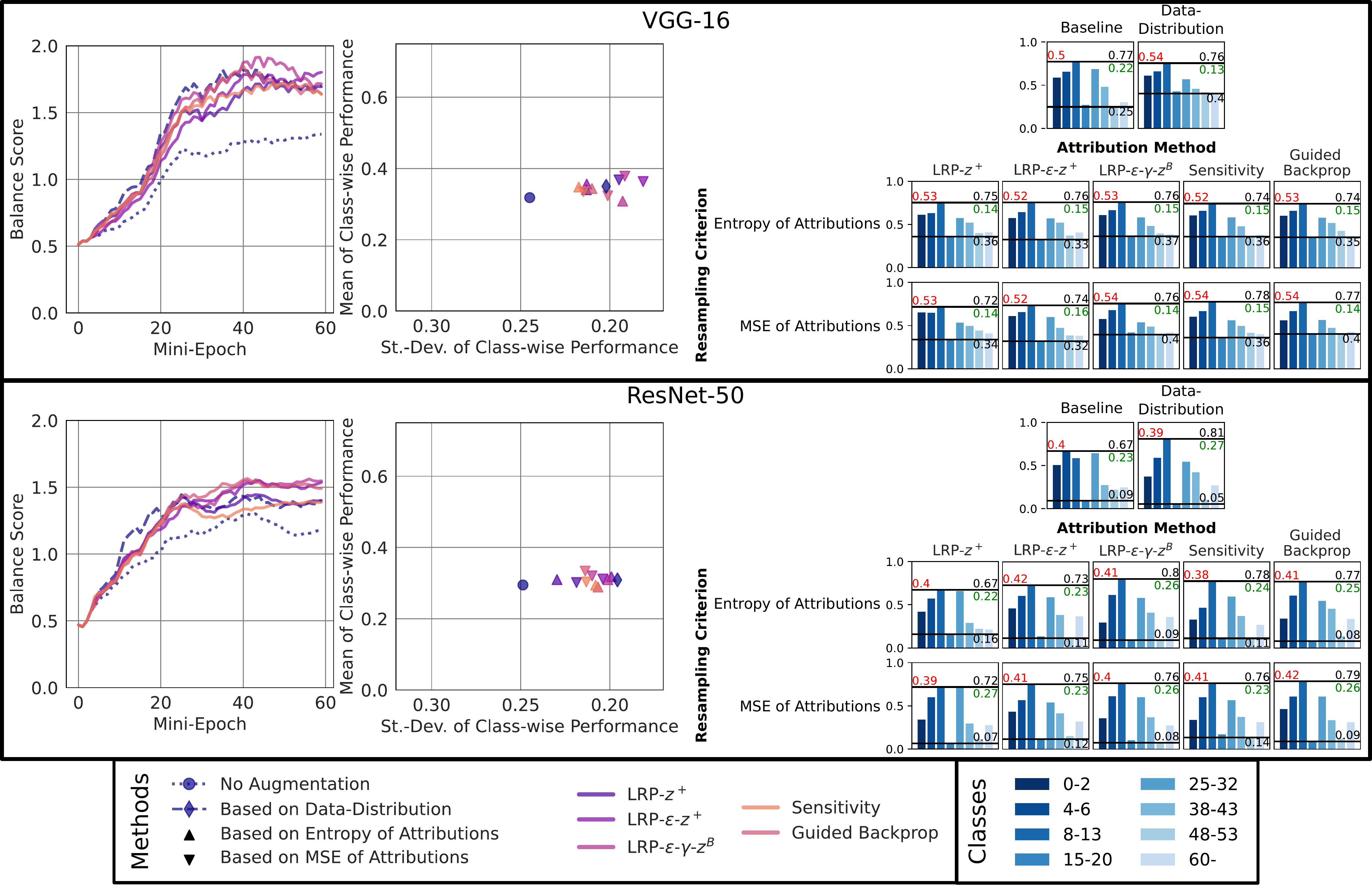}
    \caption{Redistribution of samples per mini-epoch based on attribution, compared to an unaugmented baseline and a control setting (based on class-wise datadistribution) on the Adience dataset. Results are shown for VGG-16 (\emph{top}) and ResNet-50 (\emph{bottom}). In terms of performance, all rebalancing techniques achieve a higher balance score (\emph{left} of each panel), as well as a lower standard deviation (\emph{center} of each panel, note that smaller values are to the right on the standard deviation axis), implying a more equalized class-wise performance. The predicted probabilities for the true class label are shown to the \emph{right}. Here, minimum and maximum values are indicated by the \emph{black} lines and numbers. The mean over all classes is shown in \emph{red}, and the average difference between classes in \emph{green}. For VGG-16, the predicted probability of the true class becomes more balanced in all settings compared to the baseline, however, this is not the case as reliably for ResNet-50. 
    }
    \label{fig:adience-batch-resampling}
\end{figure}

\begin{figure}[!ht]
    \centering
    \includegraphics[width=\linewidth]{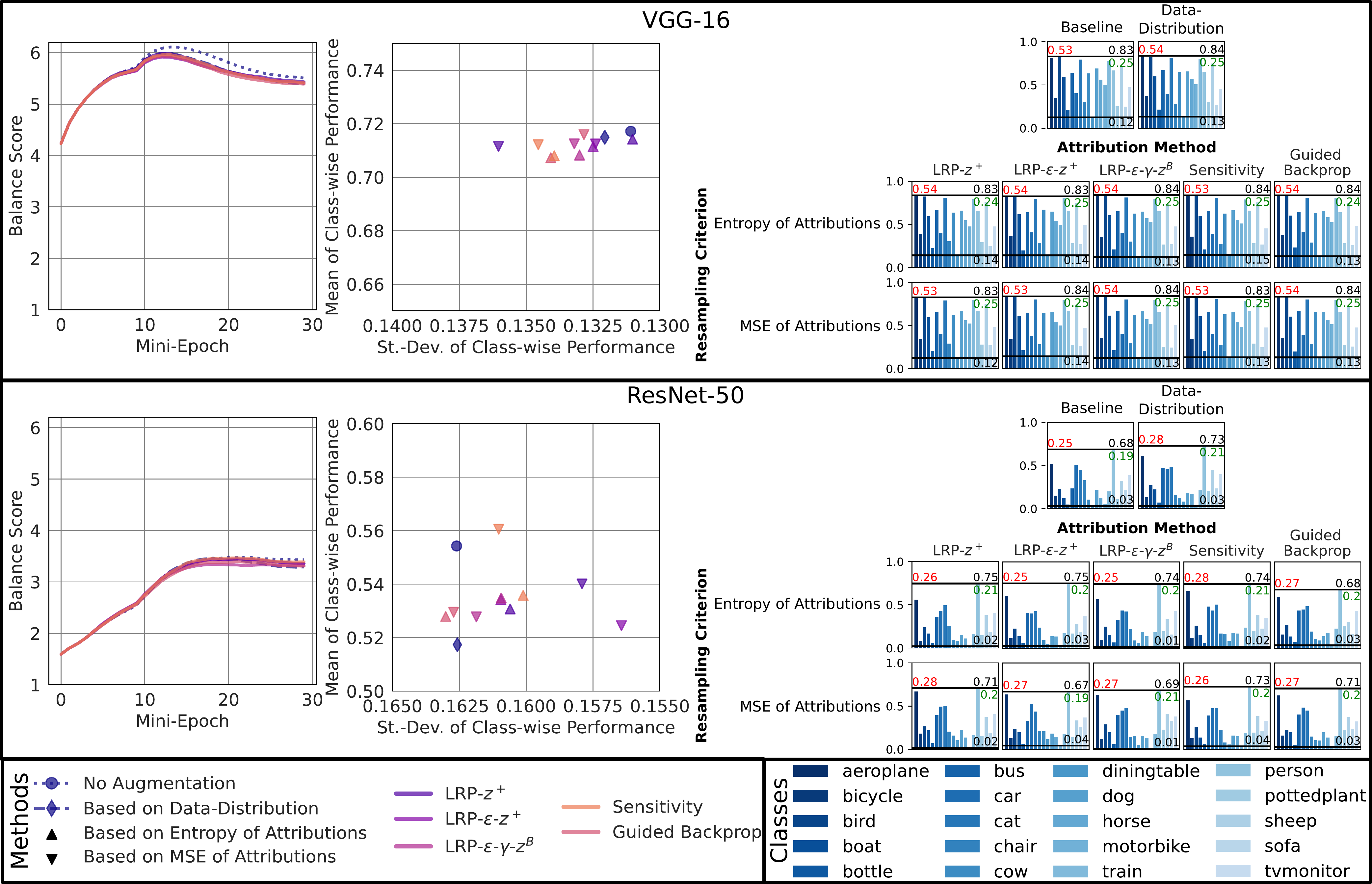}
    \caption{Redistribution of samples per mini-epoch based on attribution, compared to an unaugmented baseline and a control setting (based on class-wise datadistribution) on the Pascal-VOC dataset. Results are shown for VGG-16 (\emph{top}) and ResNet-50 (\emph{bottom}). 
    Here, model balance score (\emph{left} of each panel), and class-wise performance mean and standard deviation (\emph{center} of each panel, note that smaller values are to the right on the standard deviation axis) do not vary much between settings. In fact, the baseline seems to have a slightly more balanced performance for VGG-16 and ResNet-50.
    The predicted probabilities for the true class label are shown to the \emph{right}. Here, minimum and maximum values are indicated by the \emph{black} lines and numbers. The mean over all classes is shown in \emph{red}, and the average difference between classes in \emph{green}. However, while for VGG-16 a slight improvement in equality is indicated by the predicted true class probabilities, barely any consistent effect is visible for ResNet-50. 
    }
    \label{fig:pascalvoc-batch-resampling}
\end{figure}

Results are shown in Figures \ref{fig:adience-batch-resampling} and \ref{fig:pascalvoc-batch-resampling}.
In each of these figures, the \emph{top} panel depicts results for VGG-16 \cite{Simonyan2015Very}, while the \emph{bottom} panel shows results for ResNet-50 \cite{He2016Deep}. The balance scores $b_p$ over mini-epochs (refer to \ref{sup:sec:demodetails:equality}) are shown to the \emph{left} of each panel. Here, results are averaged over the employed resampling criteria (Entropy and MSE-distance of attributions, refer to \cite{Weber2020Towards}), and a sliding window of 10 mini-epochs was applied to reduce noisiness of the lines. Standard deviation and mean of class-wise performances of the final models are depicted in the \emph{center} of each panel. Here, a lower standard deviation and a higher mean performance are desirable, that is, models that score more towards the top right of the plot have better and more similar class-wise performance, since the x-axis is inverted. To the \emph{right} of each panel, the predicted probabilities for the true class label are shown for the final models of all investigated settings, averaged over the respective test set. Here, minimum and maximum average probabilities are shown in black (a higher minimum and lower maximum are better), and the mean over all classes in red (higher is better). The green number denotes the average difference in predicted probability between classes (lower is better).

\begin{figure}[!ht]
    \centering
    \includegraphics[width=\linewidth]{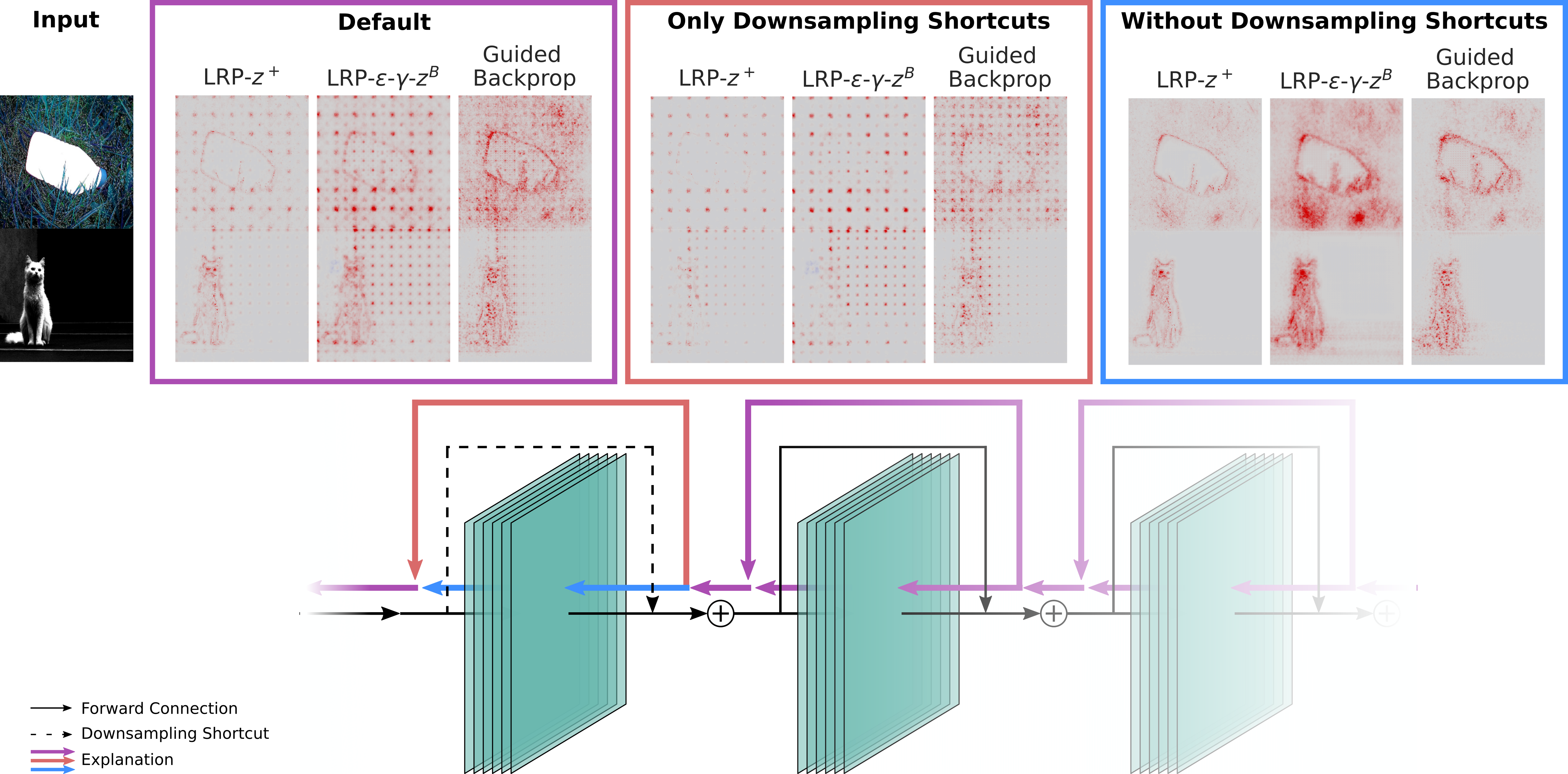}
    \caption{Examples for the downsampling artifact present when explaining decisions of ResNet-50 with various explanation methods. A visualization of the explanation flow through a segment of ResNet-50 is visualized at the \emph{bottom}. \emph{Left to Right}: 1. Example input images from the Pascal-VOC Dataset. 2. Default explanations of these samples for \gls{lrp}-$z^+$, \gls{lrp}-$\varepsilon$-$\gamma$-$z^B$, and Guided Backpropagation without excluding any paths through the model; the downsampling artifact (evenly spaced dots) is clearly visible here. 3. Partial explanations where only the downsampling shortcut paths are used for the modified backpropagation; almost only the artifact is visible here. 4. Partial explanations where the downsampling shortcut paths are excluded from the explanation computation; the downsampling artifact is gone.
    }
    \label{fig:resnet50-attribution-artifact}
\end{figure}

For the Adience dataset (see Figure~\ref{fig:adience-batch-resampling}), all augmented models achieve a higher balance score over the course of training than the unaugmented baseline (\emph{left}), both for VGG-16 and ResNet-50, although the effect is far more significant for the former. For the final models, the augmented models achieve a similar or slightly higher mean class-wise performance compared to the baseline, but a far lower standard deviation, implying increased equality (\emph{center}). Again, this effect is more pronounced for VGG-16 than for ResNet-50. Looking at the average predicted true class probabilities of the final models (\emph{left}), the baseline has the lowest minimum predicted true class probability for VGG-16. This is strongly improved upon by the redistribution based on data distribution, and by the explanation-based variants, especially \gls{lrp}-$\varepsilon$-$\gamma$-$z^B$ and Guided Backpropagation. There is, however, a notable trade-off between increasing the minimum and decreasing the maximum predicted probability. While all augmentations lead to increased minimum probabilities, most of them also reduce the maximum probability. However, all augmentation methods seem to increase the average predicted probability, while lowering the average distance between class-wise probabilities for VGG-16, showing an increase in model equality. The predicted probabilities are far more imbalanced for the ResNet-50 baseline model than for VGG-16, indicating that this model tends to overfit more in the given setting. While most augmentations increase the minimum predicted probability here, the average across classes is not increased as consistently as for VGG-16. Interestingly, the redistribution based on the data-distribution performs far worse than the baseline, as opposed to the VGG-16 setting, showing inconsistent behavior. Moreover, the average distance between class-wise predicted probabilities never decreases for ResNet-50, as opposed to VGG-16. Investigating the explanations of ResNet-50 more closely, we find that the downsampling shortcuts within ResNet-50, specifically those with kernel-size 1$\times$1 and stride 2, lead to a significant artifact for some explanation methods. Some examples for this effect (from the Pascal-VOC dataset) can be found in Figure~\ref{fig:resnet50-attribution-artifact}. While computing attributions without considering those shortcuts removes the downsampling artifact, the resulting attributions would also only partially explain the model decisions. However, this artifact may affect the employed attribution-based augmentation methods, making them perform less reliably on ResNet-50 than on VGG-16, although the performance equality is still visibly (albeit less significantly) improved (\emph{center and left}).

For the Pascal-VOC dataset (see Figure~\ref{fig:pascalvoc-batch-resampling}) the balance score is lower than baseline in all settings. Both for VGG-16 and ResNet-50, rebalancing seems to slightly lower the balance score over mini-epochs (\emph{left}). However, the differences between models are extremely small, and the balance scores are generally far higher than for the Adience dataset in Figure~\ref{fig:adience-batch-resampling}) --- even for the baseline models. Compared to Adience, the difference between classes is significantly lower for Pascal-VOC, as indicated by the generally lower standard deviations (\emph{center}) compared to Figure~\ref{fig:adience-batch-resampling}, both for VGG-16 and ResNet-15. This may be the case because the domain of Pascal-VOC is semantically closer to ILSVCR2012 \cite{Russakovsky2015Imagenet} (since all models used here were pre-trained on ILSVCR2012) than the Adience dataset is. The models therefore perform comparatively well on Pascal-VOC from the start, and do not change during training as significantly as for Adience, an effect also increased by the comparatively small learning rate ($0.05$). The final augmented models have similar or slightly lower mean class-wise performances for VGG-16 and ResNet-50, a slightly lower standard deviation than the baseline model for VGG-16, and a slightly higher one for ResNet-50 (albeit the considered value range is extremely small here, compared to the experimentes on Adience) (\emph{center}). Because samples from Pascal-VOC can contain more than one true class label due to the multi-label setting, the rebalancing of mini-batches employed here may not be as impactful as for Adience, since when adding or removing samples for one class, other classes may also be affected. However, similar to Figure~\ref{fig:adience-batch-resampling}, all augmentations seem to improve upon the baseline in terms of predicted true class probabilities for VGG-16 (higher minimum and average predicted probabilities, with lower or similar average distance between probabilities, compared to the baseline), albeit far less significantly. Interestingly, while minimum probabilities increase, maximum probabilities do not seem to decrease, which may be related either to the reduced impact of the augmentation compared to Adience or to the large amount of classes and the multilabel setting. Again, ResNet-50 seems to overfit more, and augmentations barely improve upon this, assumedly due to the attribution artifacts caused by the downsampling shortcut connections (Figure~\ref{fig:resnet50-attribution-artifact}). While the average predicted probabilities (\emph{red}) mostly increase, the average distances between probabilities also mostly increase with augmentation (\emph{green}) while the minimum probabilities mostly decrease. It seems that especially for ResNet-50, the augmentations barely have any effect on class equality in this setting.

\subsection{Lessons Learned}

The above experiments demonstrate that augmentations based on \gls{xai} are able to improve model properties such as performance or equality even in highly complex settings. In Section~\ref{sec:demoexamples:performance}, \gls{xai} was used to achieve small improvements to the test accuracy of few-shot classifiers in a consistent manner. In Section~\ref{sec:demoexamples:equality}, explanations were employed to achieve more equal class-wise performances in imbalanced settings, with varying success.
We found that in complex settings, the achieved effects may not always be as significant as in toy settings (\eg, compare Toy Experiment 1 in Section~\ref{sec:background:improving} to Section~\ref{sec:demoexamples:performance}). However, there are also often no notable drawbacks to employing \gls{xai}-based model improvement in order to, \eg, increase few-shot accuracy by a few percent (Section~\ref{sec:demoexamples:performance}). 

We further noted that \gls{xai}-based techniques can be better than the alternatives, \ie, approaches not based on \gls{xai}, but this is not always the case, as demonstrated in Section~\ref{sec:demoexamples:equality}, and should therefore not be blindly applied. Explainability methods and model improvement techniques derived from them are highly dependent on various hyperparameters, such as the employed task, dataset, model (see, \eg, Figure~\ref{fig:resnet50-attribution-artifact}), and \gls{xai}-method. These parameters therefore need to be carefully selected based on the specific setting. For instance, which \gls{xai}-method is used to obtain explanations can significantly impact the success of a model improvement technique (Sections \ref{sec:demoexamples:performance} and \ref{sec:demoexamples:equality}). 
As such, while \gls{xai}-based model improvement techniques can be useful even in complex settings, \eg, to simply achieve higher performance or to affect model properties that are non-trivial and difficult to quantify, since \gls{xai} offers detailed information about the model behavior, these methods in turn need to be applied with caution, since they are often sensitive to various hyperparameters and do not always outperform alternatives not based on \gls{xai}.

\section{Conclusion}
\label{sec:conclusion}
With the advance of explainability methods that are able to inform about the decision-making behavior of modern \gls{ml}-architectures, especially \glspl{dnn}, these methods could serve as tools for improving models beyond simple statistics such as test performance, but have barely been applied beyond explaining decisions and discovering problems in existing models. Only recently, more and more research has been published that employs \gls{xai} in practice, augmenting models in order to improve various beneficial (and often complex or intangible) properties such as model reasoning and efficiency.

In this paper, we illustrated how \gls{xai} can be used to improve models through various toy examples, and that the resulting effects can be significant and beneficial. We further introduced a formalized categorization of \gls{xai}-dependent model augmentations based on the component of the model training process they are applied at, in order to systematically review and compare approaches from existing research. Finally, we investigated the effect of \gls{xai}-based augmentations in complex settings, as well as the drawbacks, limitations, and caveats of their application. 

\gls{xai} can be practically applied to achieve various beneficial effects. In contrast to the generally utilized test performance statistics such as accuracy, precision, or recall, it offers comprehensive feedback on complex properties that can be used to diagnose and significantly improve models. Depending on how the model or the training process are altered, and what the desired goal is, various restrictions may apply. For instance, reasoning is a property that can ensure a (truly) general understanding of the data and task at hand --- unaffected by any confounders or biases within the data. While \gls{xai} can offer information about which features a model uses to make its predictions, the decision whether these features should be used cannot be derived from data or explanations alone, and requires some additional ground truth, \eg, provided by a human expert. Augmenting the loss function based on \gls{xai} is by far the most researched type of augmentation, since it allows for various improvement goals to be easily expressed as regularization terms. However, when explanations are part of these regularization terms, this restricts the number of applicable methods to gradient-based and modified backpropagation techniques, since they need to be differentiable \wrt\ model weights. Furthermore, not every type of augmentation is applicable to directly improve any property. For instance, augmenting the model after training has finished --- via pruning or quantization based on \gls{xai} --- can provide significant improvements to model efficiency, but is not suitable to improve any other properties.

However, we found that despite the various benefits of utilizing the information gained through \gls{xai} to augment and improve \gls{ml}-models, this application is not always trivial. \gls{xai} offers useful but complex information that is often extremely dependent on factors such as the employed explanation method, model, dataset, task, and hyperparameters such as normalization, and can therefore be difficult to exploit reliably. As shown by our experiments on relatively complex tasks in Section~\ref{sec:demoexamples}, effects of augmentations in these settings may be comparatively minimal or vary due to above factors. \gls{xai}-dependent augmentations may further invoke potentially unexpected side-effects \cite{Dombrowski2019Explanations, Anders2020Fairwashing}, and related techniques need to be applied carefully and with caution. Nevertheless, our experiments also showed that under the right conditions, \gls{xai} can provide significant, diverse, and reliable benefits. 
As such, practically employing explanations to improve \gls{ml}-models is not only a promising concept, but extremely helpful in improving non-trivial model properties. Nevertheless, it requires careful consideration and further research. While most approaches focus on augmenting a single component of the training process, different augmentations (\eg, data and loss function) do not interfere with each other, and combining multiple ones could provide a significant improvement to a specific property, however, this is subject to future work. Moreover, it should be noted that \gls{xai}-based augmentations often aim at improving different, more complex, and less tangible properties than test performance, such as equality or reasoning. These goals are of a semantic nature, and strict test performance in fact often needs to be sacrificed in order to achieve them. For instance, in terms of reasoning, a model that overfits on Clever Hans type of features may achieve a high test performance, but classify based on the wrong evidence. When improving reasoning, these (apparently useful) features are forbidden to the model, and test performance usually decreases as a result. However, since a model that relies on the wrong reasons cannot easily generalize to new data, it is not practically useful in real-world applications. Established test performance metrics may therefore be outdated, as different properties become more and more important, and new metrics may be required. For this purpose, \gls{xai} could be a valuable asset, as it is able to provide semantic information about model behavior, and its usefulness for improving complex model properties should therefore be explored further.

\section*{Acknowledgments}
This work was supported by the German Ministry for Education and Research (BMBF) [grant numbers 01IS14013A-E, 01GQ1115, 01GQ0850, 01IS18025A and 01IS18037A];
the iToBoS (Intelligent Total Body Scanner for Early Detection of Melanoma) project funded by the European Union’s Horizon 2020 research and innovation programme [grant agreement No 965221];
the Research Council of Norway, via the SFI Visual Intelligence grant [project grant number 309439], and UiO dScience -- Centre for Computational and Data Science.

\bibliographystyle{unsrtnat}
\bibliography{main_20220315163325}

\clearpage

\setcounter{figure}{0}
\setcounter{table}{0}
\setcounter{page}{1}
\pagenumbering{roman}
\renewcommand{\figurename}{Supplementary Figure}
\renewcommand{\tablename}{Supplementary Table}
\renewcommand\thefigure{\arabic{figure}}
\renewcommand\thetable{\arabic{table}}

\begin{appendix}

\section{Details about Toy Experiments}
\label{sup:sec:toydetails}
\gls{lrp} \cite{Bach2015Pixel} with the $\varepsilon$-$z^+$-composite (i.e., $\varepsilon$-rule for fully-connected layers and $z^+$-rule for convolutional layers), as implemented in the Zennit library \cite{Anders2021Software}, is employed to compute explanations \wrt the true class label of each sample. Each experiment was repeated five times with randomly drawn seeds and averaged over these variations in order to obtain the results shown and discussed below.\\

\subsubsection*{Toy Experiment 1 (Model Performance)}
\begin{figure}[ht]
    \centering
    \includegraphics[width=0.75\linewidth]{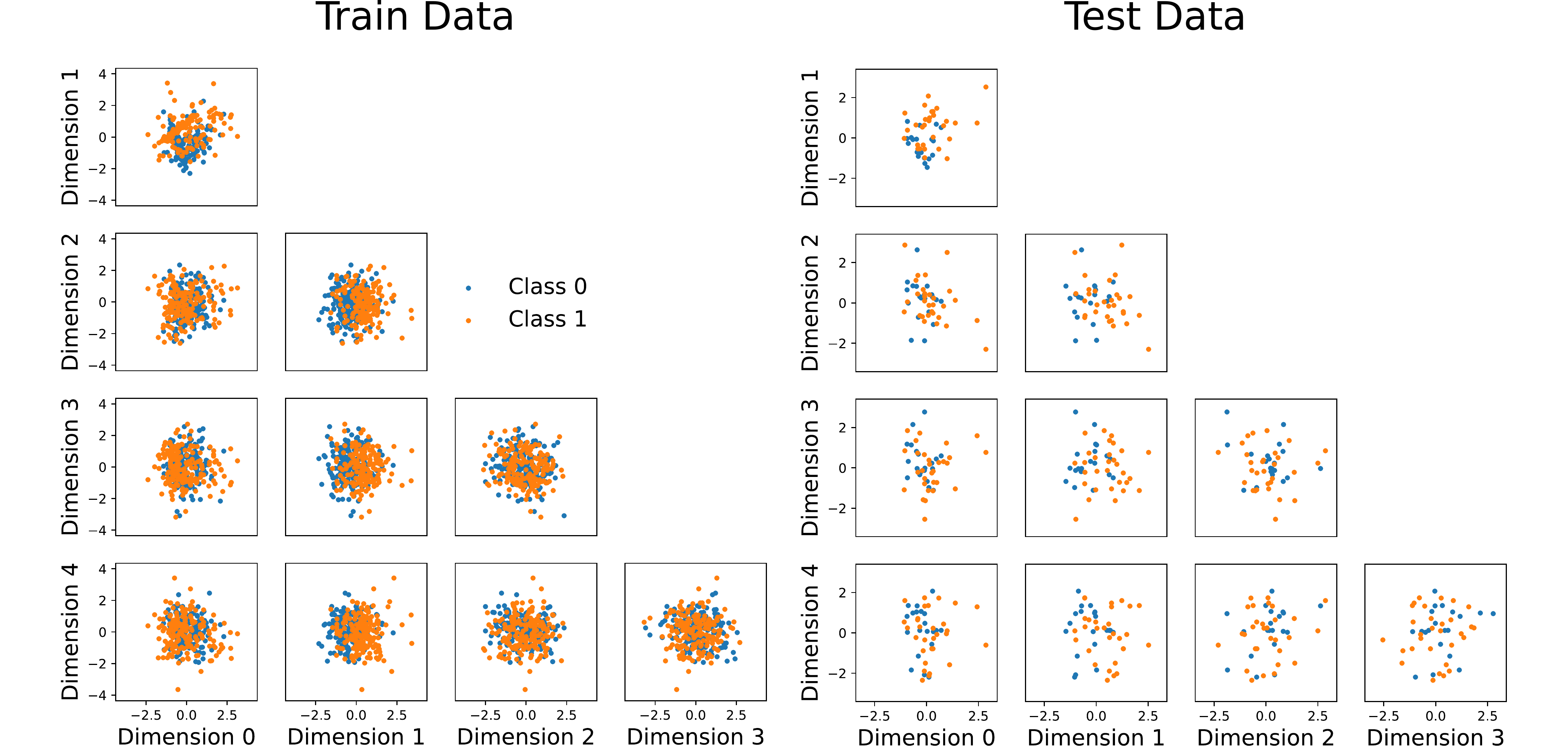}
    \caption{Visualization of the dataset used in Toy Experiment 1. The vertical axis of each panel shows the data dimension corresponding to the row index of panels, the horizontal axis shows the data dimension corresponding to column index of the grid. Training data can be found to the \emph{left}, test data to the \emph{right}.}
    \label{fig:toy-dataset-1}
\end{figure}

We first generate a  five-dimensional dataset describing a binary classification problem consisting of 400 samples, and two gaussian clusters of samples per class. The first two of these input dimensions are independent and informative while the last three consist of random noise. To make classification more difficult, 10\% of the class labels are assigned randomly. 50 samples are retained as a test set.
The data for this toy experiment is visualized in Figure \ref{fig:toy-dataset-1}.
A four layer fully-connected model (64, 32, 16, and 2 neurons), with ReLU activations after the first three layers, and softmax for the last one, is trained on this data for 500 iterations with batch-size 32, employing cross-entropy loss and an \gls{sgd} \cite{Bottou2012Stochastic} optimizer with learning rate 0.01 and a momentum of 0.9. In addition to the unaugmented baseline models, \gls{xai}-based feature augmentation is performed as follows:

Intermediate features are weighted during the forward pass. This is similar to \cite{Sun2020Explanation}, although there are some differences, \eg, in order to keep the experiment simple, we compute attributions \wrt the true class.
Given samples $x_i, i \in \{1, ..., N\}$ in a data batch of size $N$, a model $f_{\theta^t}$ with parameters $\theta^t$, and attributions $r_i^{l, t}$ \wrt intermediate features $f^l_{\theta^t}(x_i)$ at the input of layer $l$ and iteration $t \in \{1, ..., T\}$ ($T = 500$ in this experiment), we first normalize them as ${(r_i')^{l, t}} = \frac{r_i^{l, t}}{max(|r_i^{l, t}|)}$,
and then obtain a feature-wise and sample-wise attention mask as 

\begin{equation}
M_\text{feat}^{i, l, t} = 0.5 + \frac{{(r_i')^{l, t}}+1}{2},
\label{eq:toy-aug:feature}
\end{equation}
where $M_\text{feat}^{i, l, t} \in [0.5, 1.5]$. The reweighted features are then obtained as ${f^l_{\theta^t}(x_i)}' = M_\text{feat}^{i, l, t} \odot f^l_{\theta^t}(x_i)$,  with $\odot$ denoting the element-wise product. In this experiment, we set layer $l = 1$ in order to make use of an internal representation, as opposed to the input, but at the same time affect a large amount of successive layers.

Each attribution is normalized by dividing it by its maximum absolute value. The input dimension-wise attribution values over iterations are depicted in Figure \ref{fig:toy-experiments}A (\emph{right}). To reduce visual noisiness, the mean between iterations 0 and $t$ is depicted at iteration $t$ for each feature.

\subsubsection*{Toy Experiment 2 (Model Performance)}
\begin{figure}[ht]
    \centering
    \includegraphics[width=0.75\linewidth]{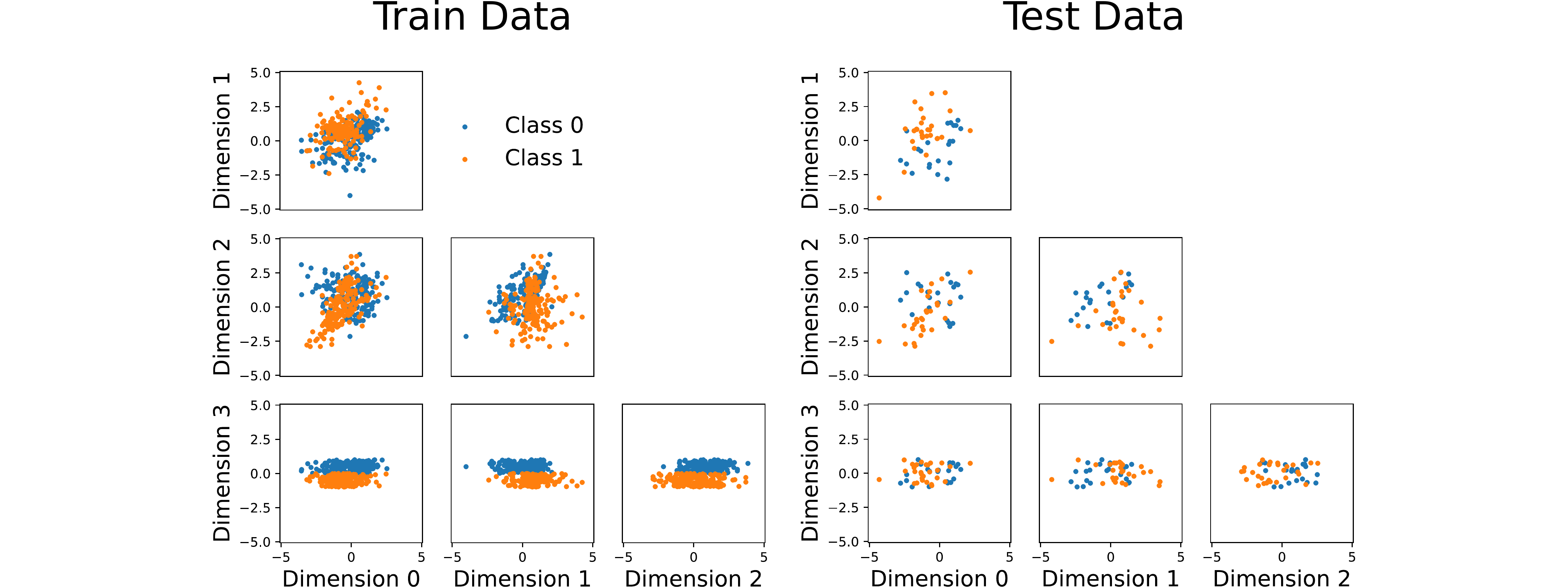}
    \caption{Visualization of the dataset used in Toy Experiment 2. The vertical axis of each panel shows the data dimension corresponding to the row index of panels, the horizontal axis shows the data dimension corresponding to column index of the grid. Training data can be found to the \emph{left}, test data to the \emph{right}.}
    \label{fig:toy-dataset-2}
\end{figure}

We again generate a dataset with 400 samples (350 train samples, 50 test samples) describing a binary classification problem, similar to the one used in the Toy Experiment 1. Here, however, each sample only contains four dimensions, with dimensions 0--2 being (truly) informative, and dimension 3 indicating the correct class via its sign for the training samples, but being randomized for the test samples.
The dataset used in this toy example is visualized in Figure \ref{fig:toy-dataset-2}.
Here, we also only assign 5\% of the class labels randomly, since we are using a smaller model than in the above experiment. More concisely, this architecture only consists of two fully-connected layers with five and two neurons, as well as ReLU and softmax activations, respectively. All models are trained for 200 iterations with batch-size 50, again using a cross-entropy loss function and \gls{sgd} \cite{Bottou2012Stochastic} optimizer with learning rate 0.01 and a momentum of 0.9 each.

A generalizing model would not only rely on the distractor dimension 3 for its predictions, but instead leverage information from all informative input dimensions --- \ie, all four dimensions for this experiment. In order to improve upon this generalization, we propose an explanation-guided dropout method, similar to \cite{Zunino2021Excitation}, which temporarily turns off the intermediate features that the model uses most to make its predictions: For this technique, we first compute attributions $r_i^{l, t}$, similar to the previous experiment. The absolute value of these is then taken in order to measure feature \emph{importance} and is then normalized, so that ${(r_i')^{l, t}} = \frac{|r_i^{l, t}|}{max(|r_i^{l, t}|)}$. At iteration $t$, the $p\%$ neurons of layer $l-1$ that correspond to the largest values of ${(r_i')^{l, t}}$ are dropped out, while the activations of all other neurons in layer $l-1$ are rescaled in order to preserve the total average activation.

To measure the effect of explanation-guided dropout on overfitting, we compare unaugmented baseline models against models that employ random (\ie, the standard) dropout and models that employ explanation-guided dropout instead. We choose a dropout rate of $p = 25\%$, since one fourth of input dimensions are distractors, and layer $l = 1$.

To obtain the graph in Figure \ref{fig:toy-experiments}B (\emph{right}), the attribution of each sample is first normalized by its respective maximum absolute value, in order to portray the relative importance of each input dimension.
Especially in the graph on the \emph{right}, this becomes apparent, although the relative relevance of the explanation-guided dropout models still grows over iterations, but much slower than for the other models, confirming the insights gained from the loss graphs.

\subsubsection*{Toy Experiment 3 (Model Reasoning)}
\begin{figure}[ht]
    \centering
    \includegraphics[width=0.75\linewidth]{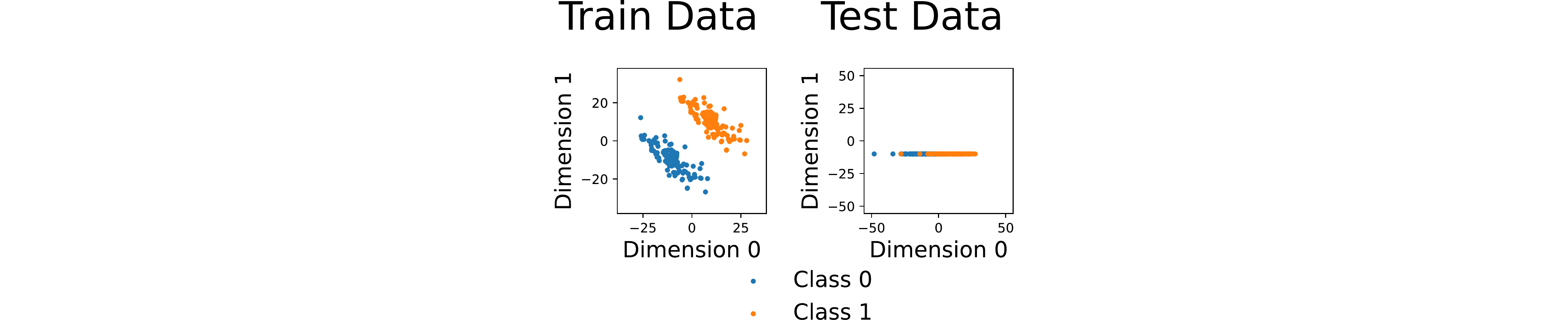}
    \caption{Visualization of the dataset used in Toy Experiment 3. The vertical axis of each panel shows the data dimension corresponding to the row index of panels, the horizontal axis shows the data dimension corresponding to column index of the grid. Training data can be found to the \emph{left}, test data to the \emph{right}.}
    \label{fig:toy-dataset-3}
\end{figure}

For this experiment, we first generate a dataset with 400 samples (200 in train and test set each) with two features, describing a binary classification problem. The data is visualized in Figures \ref{fig:toy-dataset-3}  and \ref{fig:toy-experiments}C (\emph{middle}).
In Figure \ref{fig:toy-experiments}C (\emph{middle}), the training set consists of the pastel colored points, and the test set is shown in saturated colors. While the training set varies in the direction of feature 1, the test set does not. On this data, we train a model consisting of a single fully connected layer with two neurons and a softmax activation function, for 200 iterations with batch-size 50. We employ a crossentropy loss function to measure the prediction error, and an \gls{sgd} \cite{Bottou2012Stochastic} optimizer with learning rate 0.001 and momentum 0.9.

Given samples $x_i$ and labels $y_i$, we first compute corresponding attributions $r_i^{l, t}$, and normalize them as ${(r_i')^{l, t}}' = \frac{|r_i^{l, t}|}{max(|r_i^{l, t}|)}$. we augment the loss function similar to \cite{Ross2017Right} as 

\begin{align}
& \mathcal{L}_{\text{loss-aug}}^{l, t}(x_i) = \mathcal{L}_{\text{pred}}(f_{\theta^t}(x_i), y_i) + \mathcal{L}_{\text{reason}}(r_i, r_A), \\
\text{with } & \mathcal{L}_{\text{reason}}(r_i, r_A) = (||(1 - r_A) \odot {(r_i')^{l, t}}||_2)^2,
\label{eq:toy-aug:loss}
\end{align}

where $\mathcal{L}_\text{pred}$ is the standard classification loss (here, cross-entropy), $||\cdot||_1$ the $\ell_1$-norm, and $r_A$ a binary ground truth mask (equal to 1 for important input dimensions). $r_A$ is turned into the corresponding \emph{irrelevancy} mask (which is employed, \eg, by the authors of \cite{Ross2017Right}), by subtracting it from 1. In contrast to \cite{Ross2017Right}, due to the simplicity of the setting, we do not consider ground truth mask on a \emph{per-sample} basis, but the same one for the whole dataset instead. Through the regularization term, the model is rewarded for aligning its explanations with the ground truth explanations. For this experiment, we computed input explanations before layer $l = 0$ and chose $r_A = (1, 0)^\top$ to focus on input dimension 0 and ignore dimension 1. 

In Figure \ref{fig:toy-experiments}C (\emph{right}), the absolute value of each attribution was first taken, and each attribution was then normalized by its respective maximum value, in order to compare the relative importance of each input dimension. 

\section{Details about Demonstrative Examples}
\label{sup:sec:demodetails}

\subsection{Example 1 (Model Performance)}
\label{sup:sec:demodetails:performance}
These experiments extended upon \cite{Sun2020Explanation}. Refer to this work for further details \wrt models, data, and implementation.

\subsection{Example 2 (Model Equality)}
\label{sup:sec:demodetails:equality}

For these experiments, we trained the pretrained VGG-16 \cite{Simonyan2015Very} (with batch normalization) and ResNet-50 \cite{He2016Deep} models available from the PyTorch \cite{Paszke2019Pytorch} model zoo, on the Adience and Pascal-VOC datasets. Adience does not contain a dedicated train set, but five data folds instead, however, for our experiments we used fold 0 for testing, and folds 1--4 as a train set. For training, we utilized a crossentropy loss and \gls{sgd} optimizer with a momentum of $0.9$. We chose a learning rate of $0.05$ for training on Pascal-VOC, while for Adience we used learning rates $0.1$ (VGG-16) and $0.01$ (ResNet-50). With Adience, we employed a learning rate decay by multiplying the learning rate with a factor of $0.3$ every 5000 iterations, and a learning rate warmup over the first 1000 iterations. For both datasets, after each sample was first normalized using the ImageNet \cite{Russakovsky2015Imagenet} mean and standard deviation, and resized to 256 x 256 pixels. After choosing a random crop of 227 x 227 pixel size and mirroring it randomly for each sample, models were trained using a batch-size of 32 for a certain number of \emph{mini-epochs}. Where an epoch normally includes all samples in the training set, a mini-epoch only contains a subset of training samples, in order to compute current attributions and employ \gls{xai}-based changes with a controllable frequency. In our experiments, we trained 60 mini-epoch of 10000 samples each for Adience and 30 mini-epochs of 10000 samples each for Pascal-VOC. 

Class-wise performance on the Adience dataset was evaluated using class-wise accuracy on a test set containing four corner-crops and one center-crop and their vertically mirrored versions (refer to \cite{Levi2015Age} for this oversampling scheme). The predicted probabilities of all 10 crops per test sample were averaged to yield the final predicted probabilities of the sample.
On Pascal-VOC, class-wise performances were evaluated by measuring the \gls{ap} per class.
We evaluate the imbalanced performance of a model by computing the mean performance $\mu_p$ over classes and seeds (3 randomly chosen seeds were evaluated for each experiment), as well as the mean standard deviation $\sigma_p$ of class-wise performance over seeds. Here, a high $\mu_p$ while maintaining a low $\sigma_p$ indicates a more balanced class-wise performance. A balance score can be computed from these as $b_p = \frac{\mu_p}{\sigma_p}$. We furthermore investigate the mean predicted probabilities of the respective true class over all test set samples, and how their class-wise distributions change through the applied augmentations. 

For computing attributions during training, we chose 20 representative samples per class randomly from the training set (160 total for Adience, 400 total for Pascal-VOC). These stayed constant over training. After each mini-epoch, we computed explanations from several local modified backpropagation techniques (since sampling-based techniques take too long to compute after each mini-epoch) for these representative samples using the Zennit library, \ie, (absolute) Sensitivity \cite{Baehrens2010How}, (absolute) Guided Backpropagation \cite{Springenberg2014Striving}, and \gls{lrp} \cite{Bach2015Pixel} with the following three composites:
(1) The \gls{lrp}-$z^+$ rule for all layers (called \gls{lrp}-$z^+$ from here on), (2) the \gls{lrp}-$z^+$ rule for all convolutional layers and the \gls{lrp}-$\varepsilon$ rule for fully-connected layers (called \gls{lrp}-$\varepsilon$-$z^+$ from here on), and (3) the \gls{lrp}-$z^B$ rule for the first layer, the \gls{lrp}-$\gamma$ rule for all convolutional layers, and the \gls{lrp}-$\varepsilon$ rule for fully-connected layers (called \gls{lrp}-$\varepsilon$-$\gamma$-$z^B$ from here on). Refer to \cite{Montavon2017Explaining, Montavon2019Layer} regarding details on the mentioned \gls{lrp}-rules. After computation, all attributions were normalized by their maximum absolute value. We did not consider attribution metrics based on sampling or surrogate models due to their low efficiency in terms of computation time, and the requirement of the employed method to compute attributions periodically.

During training, we applied the following augmentations with the aim of achieving a more balanced performance between classes:
We re-distributed the samples for each mini-epoch, \ie, samples for one mini-epoch were chosen (randomly) from all available samples, such that a portion $p_c$ of samples belong to class $c$. (I) Apart from an unaugmented baseline, (II) $p_c$ was computed by applying a softmax to the (inverted) class distribution of the whole training set. (III) Furthermore, at each mini-epoch either the entropy (see \cite{Weber2020Towards} for details) of the attributions, or the MSE-Distance to the corresponding attributions from the previous mini-epoch (see \cite{Weber2020Towards} for details) was computed for each representative sample, and then averaged class-wise. To avoid large variations, attributions were averaged over the last 5 mini-epochs before computing above measurements on them. Afterwards, $p_c$ was computed by applying softmax to these class-wise averaged measurements. Since we augmented the input data distribution, we employed input-space explanations here.

\clearpage
\section{Supplementary Toy Experiment Figures}
\label{sup:sec:toyfig}

\end{appendix}
\end{document}